\documentclass{article}

\usepackage{microtype}
\usepackage{graphicx}
\usepackage{subcaption}
\usepackage{booktabs} 

\usepackage{hyperref}
\usepackage{xurl}

\usepackage{amsmath}
\usepackage{amssymb}
\usepackage{mathtools}
\usepackage{amsthm}

\usepackage[capitalize,noabbrev]{cleveref}

\usepackage[T1]{fontenc}
\usepackage[utf8]{inputenc}
\usepackage{tcolorbox}
\tcbuselibrary{listings, breakable}

\newtcblisting{modelbox}{
  breakable,
  colback=gray!5,
  colframe=gray!30,
  listing only,
  sharp corners,
  fontupper=\small\ttfamily,
  listing options={
    basicstyle=\small\ttfamily,
    breaklines=true,                
    breakatwhitespace=true,         
    columns=flexible,
    tabsize=2,
    literate={⁹}{{$^9$}}1 {–}{{-}}1 {—}{{-}}1 { }{{\ }}1 
  }
}

\usepackage{longtable}
\usepackage{pifont} 
\usepackage{booktabs} 

\newcommand{\cmark}{\ding{51}} 

\theoremstyle{plain}

\theoremstyle{definition}

\theoremstyle{remark}

\newif\ifaccepted
\acceptedtrue 

\ifaccepted
    \usepackage[preprint]{icml2026}
\else
    \usepackage{icml2026}
\fi

\newcommand{\alttext}[2]{\ifaccepted#2\else#1\fi}

\newcommand{\onlyaccepted}[1]{\ifaccepted#1\fi}

\newcommand{\onlyblind}[1]{\ifaccepted\else#1\fi}

\usepackage[textsize=tiny]{todonotes}

\icmltitlerunning{TwinWeaver: An LLM-Based Foundation Model Framework for Pan-Cancer Digital Twins}

\begin{document}

\twocolumn[
  \icmltitle{TwinWeaver: An LLM-Based Foundation Model Framework\\for Pan-Cancer Digital Twins}



  \icmlsetsymbol{super}{+}

  \begin{icmlauthorlist}
    \icmlauthor{Nikita Makarov}{roche_de,helmholtz,lmu}
    \icmlauthor{Maria Bordukova}{roche_de,helmholtz,lmu}
    \icmlauthor{Lena Voith von Voithenberg}{roche_ch_dev}
    \icmlauthor{Estrella Pivel-Villanueva}{roche_de}
    \icmlauthor{Sabrina Mielke}{gen_ny}
    \icmlauthor{Jonathan Wickes}{gen_sf}
    \icmlauthor{Hanchen Wang}{gen_sf,stanford}
    \icmlauthor{Mingyu Derek Ma}{gen_ny}
    \icmlauthor{Keunwoo Choi}{gen_ny}
    \icmlauthor{Kyunghyun Cho}{gen_ny,nyu}
    \icmlauthor{Stephen Ra}{gen_ny}
    \icmlauthor{Raul Rodriguez-Esteban}{roche_ch_basel,super}
    \icmlauthor{Fabian Schmich}{roche_de,super}
    \icmlauthor{Michael Menden}{helmholtz,unimelb,super}
  \end{icmlauthorlist}

  \icmlaffiliation{roche_de}{Computational Sciences Center of Excellence, Roche, Penzberg, Germany}
  \icmlaffiliation{helmholtz}{Computational Health Center, Helmholtz Munich, Munich, Germany}
  \icmlaffiliation{lmu}{Department of Biology, Ludwig Maximilian University of Munich, Munich, Germany}
  \icmlaffiliation{roche_ch_dev}{Early Development Oncology, Roche Innovation Center Zurich, Roche, Schlieren, Switzerland}
  \icmlaffiliation{gen_ny}{Computational Sciences Center of Excellence, Genentech, New York City, USA}
  \icmlaffiliation{gen_sf}{Computational Sciences Center of Excellence, Genentech, South San Francisco, USA}
  \icmlaffiliation{nyu}{Center for Data Science, New York University, New York City, USA}
  \icmlaffiliation{stanford}{Department of Computer Science, Stanford University, Stanford, CA, USA}
  \icmlaffiliation{roche_ch_basel}{Computational Sciences Center of Excellence, Roche, Basel, Switzerland}
  \icmlaffiliation{unimelb}{Department of Biochemistry and Pharmacology, Bio21 Molecular Science and Biotechnology Institute, The University of Melbourne, Melbourne, Australia}

  \icmlcorrespondingauthor{Fabian Schmich}{fabian.schmich@roche.com}
  \icmlcorrespondingauthor{Michael Menden}{michael.menden@unimelb.edu.au}

  \icmlkeywords{Machine Learning, ICML}

\onlyaccepted{
\vskip 0.2in

  \begin{center}
    \parbox{0.9\textwidth}{
  \begin{abstract}
  \vskip 0.1in
Precision oncology requires forecasting clinical events and trajectories, yet modeling sparse, multi-modal clinical time series remains a critical challenge.
We introduce TwinWeaver, an open-source framework that serializes longitudinal patient histories into text, enabling unified event prediction as well as forecasting with large language models, and use it to build Genie Digital Twin (GDT) on 93,054 patients across 20 cancer types.
In benchmarks, GDT significantly reduces forecasting error, achieving a median Mean Absolute Scaled Error (MASE) of 0.87 compared to 0.97 for the strongest time-series baseline ($p<0.001$). Furthermore, GDT improves risk stratification, achieving an average concordance index (C-index) of 0.703 across survival, progression, and therapy switching tasks, surpassing the best baseline of 0.662. GDT also generalizes to out-of-distribution clinical trials, matching trained baselines at zero-shot and surpassing them with fine-tuning, achieving a median MASE of 0.75–0.88 and outperforming the strongest baseline in event prediction with an average C-index of 0.672 versus 0.648. Finally, TwinWeaver enables an interpretable clinical reasoning extension, providing a scalable and transparent foundation for longitudinal clinical modeling.
\end{abstract}
}
\end{center}
}
  \vskip 0.2in
  
]



\printAffiliationsAndNotice{\textsuperscript{+}Equal supervision}

\onlyblind{
\begin{abstract}
Precision oncology requires forecasting clinical events and trajectories, yet modeling sparse, multi-modal clinical time series remains a critical challenge.
We introduce TwinWeaver, an open-source framework that serializes longitudinal patient histories into text, enabling unified event prediction as well as forecasting with large language models, and use it to build Genie Digital Twin (GDT) on 93,054 patients across 20 cancer types.
In benchmarks, GDT significantly reduces forecasting errors, achieving a median Mean Absolute Scaled Error (MASE) of 0.87 compared to 0.97 for the strongest time-series baseline ($p<0.001$). Furthermore, GDT improves risk stratification, achieving an average concordance index (C-index) of 0.703 across survival, progression, and therapy switching tasks, surpassing the best baseline of 0.662. GDT also generalizes to out-of-distribution clinical trials, matching trained baselines at zero-shot and surpassing them with fine-tuning, achieving a median MASE of 0.75–0.88 and outperforming the strongest baseline in event prediction with an average C-index of 0.672 versus 0.648. Finally, TwinWeaver enables an interpretable clinical reasoning extension, providing a scalable and transparent foundation for longitudinal clinical modeling.
\end{abstract}
}

\section{Introduction}

\begin{figure*}
    \centering
    \includegraphics[width=\linewidth]{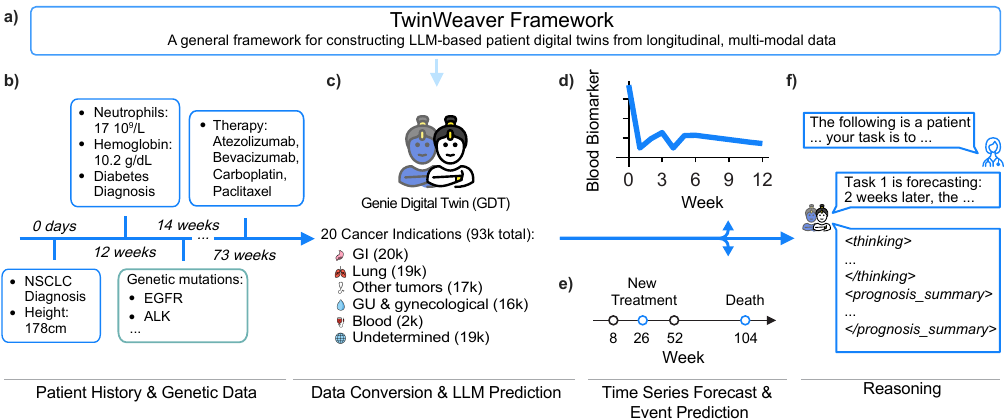}
    \caption{
    \textbf{The TwinWeaver framework serializes longitudinal patient histories into text to train the Genie Digital Twin (GDT) pan-cancer foundation model.}
    \textbf{a)} The TwinWeaver framework \textbf{b)} serializes multi-modal EHR data (history, genetics) into text. \textbf{c)} We use this to develop GDT on 93,054 real-world patients across 20 cancer indications (counts shown in parentheses). GDT jointly \textbf{d)} forecasts continuous biomarkers and \textbf{e)} predicts landmark clinical events. \textbf{f)} The model can be further trained for interpretable predictions using reasoning-based outputs.}
    \label{fig:introduction:overview}
\end{figure*}


Precision oncology aims to personalize cancer care through computational modeling of patient trajectories \cite{zhuang.2025.longitudinal_oncology}. Patient Digital Twins (PDTs) serve as digital replicas of patients with a bidirectional information flow, learning patient-specific representations from longitudinal, multi-modal molecular and clinical data to support forecasting and simulation. This necessitates scalable methods that can handle heterogeneous and sparse real-world time series \cite{kamel.2021.digital_twin, bordukova.2024.generative}. Recent advances in foundation models have further increased the feasibility of large-scale clinical trajectory modeling \cite{makarov.2025.llms, wornow.2023.shaky_foundations}.

Here, we introduce TwinWeaver, a longitudinal framework for LLM-generated PDTs, and Genie Digital Twin (GDT), a pan-cancer model instantiated using TwinWeaver to simulate patient trajectories (Figure \ref{fig:introduction:overview}). TwinWeaver serializes patient journeys into text, which GDT uses to forecast clinical events and trajectories across 20 cancer indications. 

Leveraging a large-scale dataset of 93,054 real-world oncology patients from the Flatiron Health-Foundation Medicine Clinico-Genomic Database (FH-FMI CGDB; Appendix \ref{app:datssets:formal}), we present four key contributions: 1) the open-source TwinWeaver framework\onlyaccepted{\footnote{\href{http://github.com/MendenLab/TwinWeaver}{http://github.com/MendenLab/TwinWeaver}}} for serializing clinical histories; 2) the GDT model, built using TwinWeaver, which demonstrates competitive performance relative to advanced time-series and electronic health record models; 3) demonstration of zero-shot generalization to clinical trials; and 4) highlighting a reinforcement learning-based interpretable clinical reasoning extension to the base model.

\section{Related Work}

\paragraph{Clinical Event Prediction} 

Clinical event prediction estimates time-to-event outcomes from longitudinal data using classical methods such as Cox Proportional Hazards models \cite{cox.1972.phmodel}, Random Survival Forests  \cite{ishwaran.2008.rsf}, and deep survival models \cite{katzman.2018.deepsurv}. These approaches often require manual data preprocessing to handle real-world data (RWD) and typically decouple event prediction from trajectory forecasting, limiting information gain. These limitations call for flexible, unified models that jointly capture longitudinal dynamics and censored clinical events.

\paragraph{Clinical \& EHR Foundation Models} 
Recent foundation models have shifted clinical modeling from static risk prediction toward dynamic trajectory forecasting \cite{guo.2023.ehr_fms}. LLM-based approaches such as DT-GPT \cite{makarov.2025.llms} and EHR2Path \cite{pellegrini.2025.ehr2path} exemplify this progress, yet they remain constrained by the number of forecastable variables or their temporal scope. EHR foundation models such as CLMBR-T \cite{wornow.2023.clmbr} and MOTOR \cite{steinberg.2023.motor} improve event prediction, while SCOPE \cite{hussain.2024.scope} further enables joint longitudinal and event tasks. However, these models rely on fixed, code-based vocabularies that hinder adaptability \cite{bedi.2026.zero_shot}. These constraints underscore the need for more flexible and scalable models for heterogeneous, long-horizon clinical trajectories.

\paragraph{Time-Series Forecasting Foundation Models} 
Deep learning architectures such as TiDE \cite{das.2023.tide} and foundation models such as Chronos \cite{ansari.2024.chronos, hoo.2024.tabpfn} have advanced time-series forecasting. Whereas LLMs can be used for forecasting by treating numerical sequences as text \cite{gruver.2023.llmtime}, their EHR applications remain limited to small variable sets. Moreover, general models such as Chronos capture temporal dependencies but lack medical context and struggle with integrating sparse genetic data. The limitations highlight the need for richer multi-modal temporal representations.

\paragraph{Reasoning in Clinical AI} 
Recent work, such as OncoReason \cite{hemadri.2025.oncoreason}, demonstrates the utility of LLMs for complex clinical reasoning and outcome prediction. These advances highlight the opportunity for models that couple numerical forecasting with interpretable textual rationales to improve transparency, trust, and clinical usability.

\section{TwinWeaver Framework}

We introduce TwinWeaver, a framework that enables large language models (LLMs) to perform clinical prediction by jointly forecasting numeric time-series and censored clinical events. TwinWeaver fine-tunes LLMs on procedurally generated text representations of longitudinal patient journeys, providing a flexible representation for multi-modal clinical data. This serialization leverages the tokenization and generative capabilities of LLMs to model irregular clinical timelines and learn shared latent patient states that support both continuous biomarker forecasting and discrete clinical event prediction.

We define our core notation for the model. Let $\mathcal{P}$ be the set of patients, where each $p = (\mathbf{s}_p, \mathcal{H}_p) \in \mathcal{P}$ has static attributes $\mathbf{s}_p$ and an observed history $\mathcal{H}_p = [ h_{p1}, ..., h_{pn} ]$ where $h_{pi} = (\text{timestamp}, \text{event}, \text{value})$, for a total of $n$ observations. The next-token probability of the LLM is defined as $q_\theta(x_i \mid x_{1:i-1})$ for a given token sequence $x$, with $x_i$ being the $i$-th token, and parameterized by $\theta$.

\subsection{Training}

The training pipeline preprocesses longitudinal data, generates input-target pairs by splitting patient trajectories at critical time points, such as therapy start dates, converts the pairs into text prompts, and fine-tunes a pretrained LLM with them.

For the trajectory splitting for patient $p$, we define the input as the combination of static data and the history up to a split time $t$, $\mathcal{X}_p(t) = (\mathbf{s}_p, \mathcal{H}_p(t))$, with $\mathcal{H}_p(t) = [h_{pi} \mid h_{pi} \in \mathcal{H}_p, h_{pi,\text{timestamp}} \le t]$. The target $\mathcal{Y}_p(t)$ is defined as a set of future predictions of various tasks.

\subsubsection{Forecasting Task}

The first task type is time-series forecasting, used to predict frequently measured values such as blood biomarkers or vital signs.

Managing LLM context size during training and inference requires limiting the number of variables forecast per instance; accordingly, a subset of variables is sampled for each training example to ensure computational feasibility. We denote the set of output variables as $V$, with $V' \subseteq V$ being a sampled subset. This sampling is weighted proportionally to each variable's observation count and volatility (Appendix \ref{app:forecasting_sampling}).

The forecasting target $\mathcal{Y}_p^{forecast}(t^{forecast}, V')$ comprises future values for the sampled variables $V'$ up to the forecasting horizon $t^{forecast} = t + \Delta t^{forecast}$ with $\Delta t^{forecast} \in \mathbb{N}^+$. We treat unmeasured values within this window and any values of new competing events (e.g., a new line of therapy) as missing targets (Appendix \ref{app:forecasting_details}). This subsampling strategy maintains flexibility across time horizons while ensuring computational feasibility.

\subsubsection{Landmark Events Task}
The second task, predicting a patient’s event status at a future time point, is a critical clinical task, which we formulate using the landmarking framework \cite{van.2007.dynamic_landmarking} with a focus on patient risk ranking.

Our data sampling methodology enables the model to predict event status for an arbitrarily sampled future horizon (Appendix \ref{app:landmark_events_details}; Algorithm \ref{alg:landmark-sampling}). We define $E \in \mathcal{E}$ as the specific clinical event from all possible events $\mathcal{E}$, e.g. disease progression, and  $t^{event} = t + \Delta t^{event}$ as the horizon, based on $\Delta t^{event} \in \mathbb{N}^+$. The set of event labels is $Y=\{\text{occurred}, \text{not occurred}, \text{censored}\}$. The model is thus trained to predict the patient's status of event $E$ at the landmark time $t^{event}$ as one of three classes $y \in Y$.

This formulation explicitly models the observation process and competing events by treating censoring as a distinct target class. Consistent with established real-world data methodologies \cite{latimer.2014.treatment_switching}, we handle treatment switching as a censoring event, as well as patients who reach the end of their recorded trajectory without the occurrence of death, or who exceed the dataset's global cutoff date.

\begin{figure*}[h!]
    \centering
    \includegraphics[width=\linewidth]{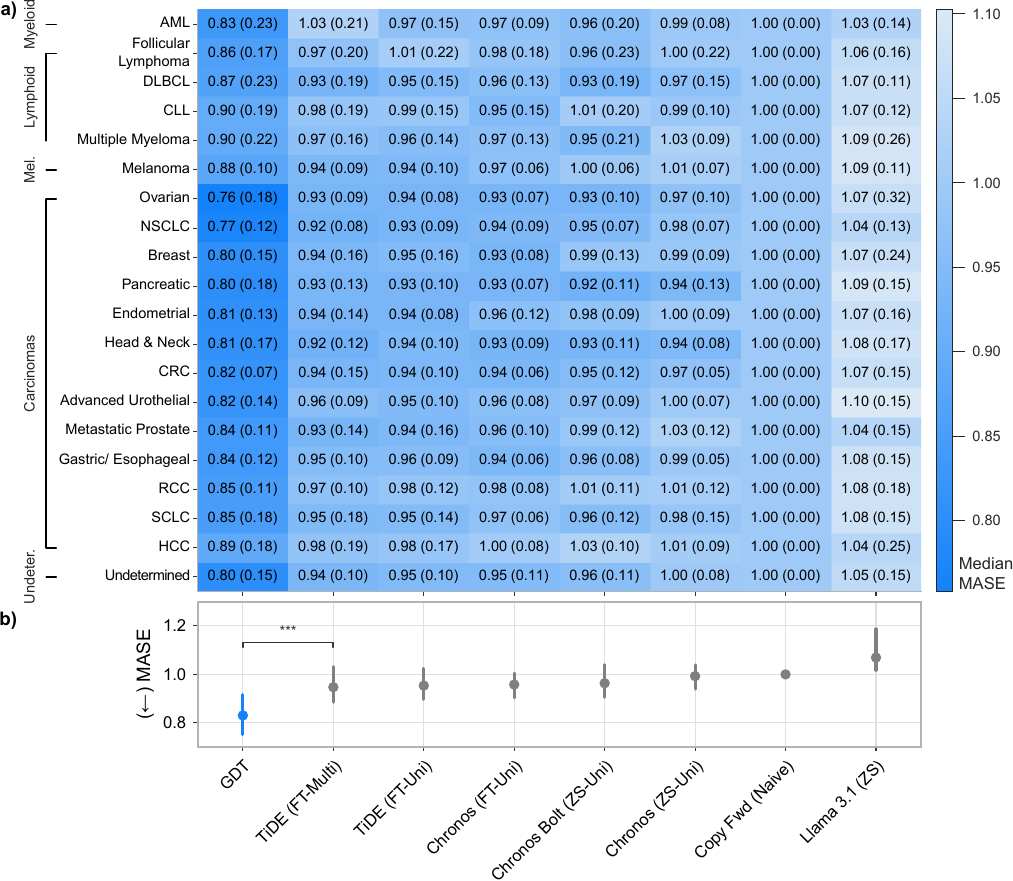}
    \caption{
    \textbf{GDT achieves reduced error in highly dynamic blood biomarker forecasting across the majority of cancer indications.} 
    \textbf{a)} Heatmap of the median Mean Absolute Scaled Error (MASE - lower is better). GDT shows lower error (darker blue) compared to baselines for the top 30 most time-changing variables per indication. \textbf{b)} Aggregated performance of the top 30 most changing variables shows GDT achieves a median MASE of 0.830, significantly outperforming the second-best baseline, TiDE multivariate ($p < 0.001$, Wilcoxon signed-rank test). Error bars and numbers in parentheses denote the Interquartile Range (IQR) across variables. ZS denotes zero-shot, FT is fine tuned, Uni is univariate input, Multi is multivariate input.}
    \label{fig:cgdb:forecasting}
\end{figure*}

\subsubsection{Combining, converting to text \& training}

The TwinWeaver framework transforms the structured patient history $\mathcal{X}_p(t)$ and multi-task targets $\mathcal{Y}_p(t)$ into a single, human-readable text prompt. As outlined in Appendix \ref{app:combining_prompt} (Algorithm \ref{alg:prompt-construction}), this serialized prompt includes static data, the chronological visit history, such as blood measurements and biomarkers, a summary of recent critical information, and a numbered list of prediction tasks. To manage context constraints, we truncate intermediate visits while strictly preserving the initial visit to retain baseline diagnostic information. Additionally, to mitigate the `lost-in-the-middle' phenomenon, we explicitly repeat the most recent clinical observations and genetic events immediately prior to the task definitions. The model is trained using a standard causal language modeling loss, computed only on the target completion tokens.

\subsection{Inference}
For evaluation, inference for forecasting and landmark event prediction is handled separately.

\subsubsection{Forecasting}
For forecasting inference, we generate $\{\hat{\mathcal{Y}_p}^{m}\}_{m=1}^M, M \in \mathbb{N}^+$ independent completions from the model's predictive distribution $q_\theta$ for a given prompt for patient $p$ \cite{makarov.2025.llms}. These $M$ text-based completions are decoded back into numerical trajectories. We then average these trajectories to produce a stable mean prediction: 

$\hat{\mathcal{Y}_p}^{\text{forecast}} = \frac{1}{M} \sum_{m=1}^M \hat{\mathcal{Y}_p}^{m}$.

\begin{figure*}
    \centering
    \includegraphics[width=\linewidth]{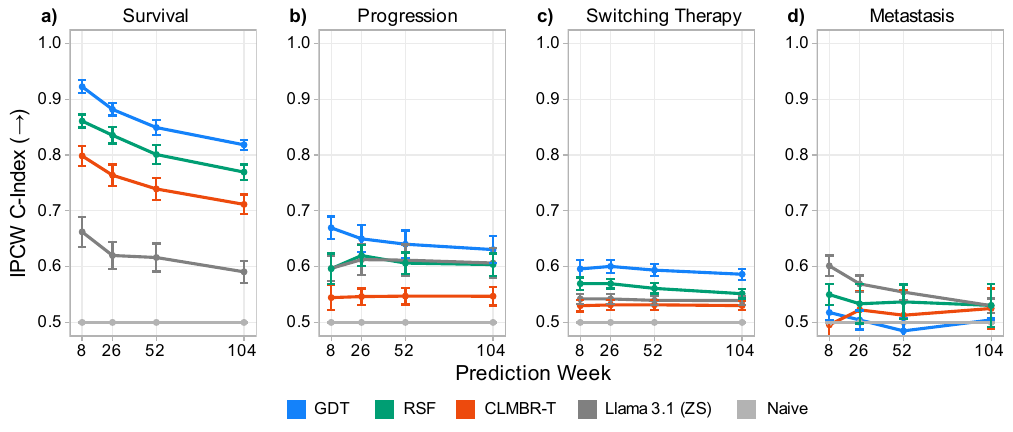}
    \caption{
    \textbf{GDT demonstrates improved risk stratification in predicting survival, progression, and switching therapy, while metastasis prediction remains constrained by limited data availability.} 
    We report the mean IPCW C-Index (higher is better) and standard error across 20 indications for \textbf{a)} survival, \textbf{b)} disease progression, \textbf{c)} switching therapy, and \textbf{d)} metastasis. GDT demonstrates robust ranking capabilities, outperforming the strongest baselines in 12 of 16 evaluated time points. Lower performance is seen in predicting metastasis outcomes, likely due to low data availability ($N=3$ indications).}
    \label{fig:cgdb:events}
\end{figure*}

\subsubsection{Landmark Events}
\label{twin_weaver:events_inference}

Event probabilities are derived from the model’s three-class outputs (occurred, not occurred, censored) using  log-likelihoods, which provide continuous scores required for survival analysis evaluation rather than discrete class predictions. Specifically, we first compute the length-normalized log-likelihood score $L_{y,p}$ for each of the three possible target strings $y \in Y$:
$$
L_{y,p} = \frac{1}{|x| - k + 1} \sum_{i=k}^{|x|} \log( q_\theta(x_i \mid x_{1:i-1}) )
$$
where $k$ is the index of the first token of the target and $x = [\text{prompt};y]$ is the prompt concatenated with the target string $y$. We then normalize these scores using a softmax function to obtain a probability estimate over the three outcomes:
$$
\hat{P}(y \mid \mathcal{X}_p(t), E, \Delta t^{event}) = \frac{\exp(L_{y,p})}{\sum_{y' \in Y} \exp(L_{y',p})}
$$
The final ranking score of the event $E$ is taken from the `occurred' class probability: 
\begin{multline*}
\text{Score}(p, E, \Delta t^{event}) = \\
\hat{P}(y=\text{occurred} \mid \mathcal{X}_p(t), E, \Delta t^{event})
\end{multline*}
\alttext{
The resulting risk score is used for all downstream evaluation metrics. Although the formulation models the joint likelihood of event occurrence and observation and lacks guarantees for monotonic cumulative incidence functions, it remains effective for \textit{ranking}-based clinical tasks such as risk stratification and comparative treatment evaluation \cite{steck.2007.ranking_good}. 

This approach preserves the LLM architecture and training procedure, enabling direct integration with existing software libraries for unified forecasting, event prediction, and reasoning. We also provide an extension with conditioning without censoring, together with monotonicity guarantees, by applying isotonic regression (Appendix \ref{app:isotonic_regression}).
}{
The resulting risk score is used for all downstream evaluation metrics. 

Although the formulation models the joint likelihood of event occurrence and observation and lacks guarantees for monotonic cumulative incidence functions, it remains effective for \textit{ranking}-based clinical tasks such as risk stratification and comparative treatment evaluation \cite{steck.2007.ranking_good}. 

This approach preserves the LLM architecture and training procedure, enabling direct integration with existing software libraries for unified forecasting, event prediction, and reasoning. We also provide an extension with conditioning without censoring, together with monotonicity guarantees, by applying isotonic regression (Appendix \ref{app:isotonic_regression}).
}

\section{Pan-Cancer LLM}

As an instance of the TwinWeaver framework, we train Genie Digital Twin (GDT), a pan-cancer LLM, on a large-scale real-world dataset from the FH-FMI CGDB (Appendix \ref{app:datssets:formal}). The cohort comprises 93,054 patients with recorded drug visits across 20 cancer types and is split at the patient level into 82,753 training, 4,991 validation, and 4,999 test patients to ensure robust evaluation.

\subsection{Data Preprocessing}

The pipeline transforms raw multi-modal inputs, including diagnoses, laboratory measurements, and genetic mutation panels, into a structured textual representation of longitudinal patient trajectories at weekly resolution (Appendix \ref{app:data_preprocessing}).

\begin{figure*}[h!t]
    \centering
    \includegraphics[width=\linewidth]{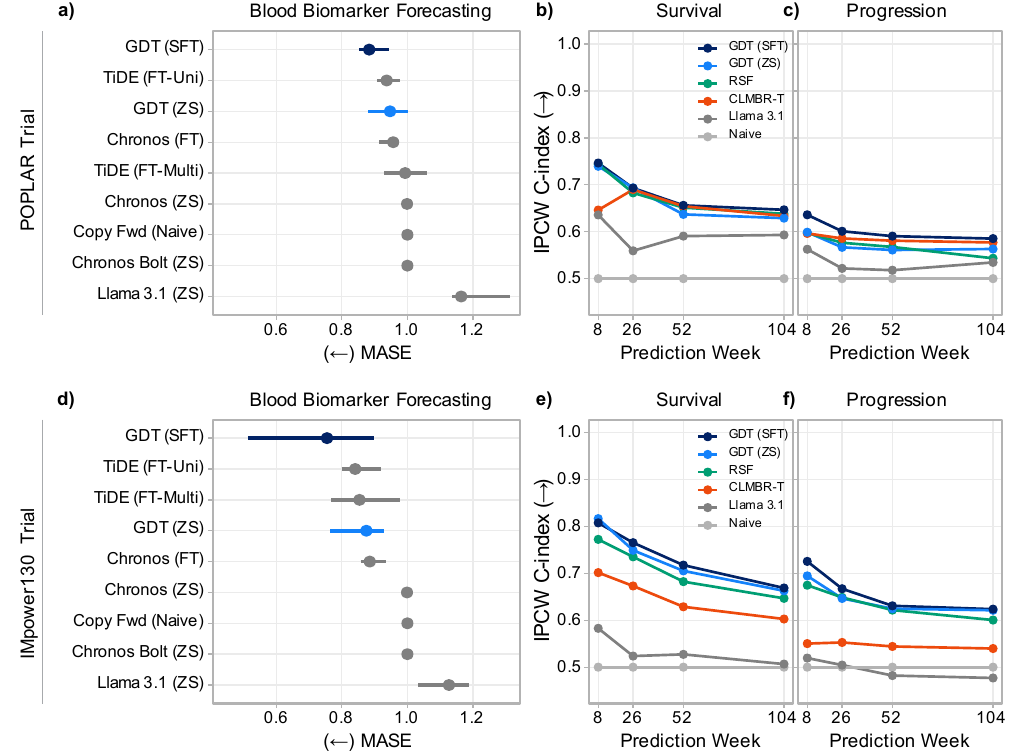}
    \caption{\textbf{GDT with supervised fine-tuning (SFT) outperforms baselines on out-of-distribution clinical trial tasks.} The model is evaluated on the unseen trials \textbf{a-c)} POPLAR and \textbf{d-f)} IMpower130. GDT (SFT) achieves the lowest MASE (Mean Absolute Scaled Error - lower is better; IQR error bars across 17 variables) in blood biomarker forecasting (\textbf{a, d}) and highest C-Indexes (higher is better) for survival (\textbf{b, e}) and progression (\textbf{c, f}) events, demonstrating strong generalization in cold start scenarios where historical data is sparse. ZS denotes zero-shot predictions, FT means the model is fine-tuned, whilst Uni is univariate input, Multi is multivariate input, RSF is Random Survival Forest.}
    \label{fig:cit:overall}
\end{figure*}

\subsection{Task Configuration}

Training focuses on clinically relevant periods by sampling split times $t$ uniformly from visits within 90 days of a new line of therapy. This strategy allows us to generate multiple training samples from a single patient without overfitting on specific patterns, specifically 10 for GDT, leveraging their full history across different treatments.

GDT is configured to perform two primary prediction tasks simultaneously. First, we set the blood biomarker time-series horizon $\Delta t^{forecast} = 13 \text{ weeks}$, a horizon highly relevant for monitoring treatment response and toxicity. Second, we predict four key clinical events (survival, metastasis, disease progression, and therapy switching) at longer horizons sampled uniformly $\Delta t^{event} \in \{1, \dots, 104\}$, enabling flexible predictions across various clinically relevant horizons. Whilst the TwinWeaver framework is model-agnostic, we choose to instantiate GDT using an established open-source model, Llama 3.1 8B Instruct. We train for 1 epoch using the AdamW optimizer, with a learning rate of $10^{-5}$ and a context length of 8000 tokens (Appendix \ref{app:training_details}).

\subsection{Reasoning Extension}

Using fine-tuned LLMs as a prediction platform enables downstream tasks such as reasoning. We focus on neutrophil forecasting on a subsample of the NSCLC cohort ($N=2,385$). To address the scarcity of gold-standard clinical rationales, we employ a knowledge distillation approach, prompting a teacher model, Qwen3 Next 80B-A3B \cite{yang.2025.qwen3}, to synthesize structured chains-of-thought conditioned on patient history and ground-truth outcomes. The student GDT model first undergoes supervised fine-tuning on this synthetic corpus, followed by alignment via Group Relative Policy Optimization (GRPO) \cite{shao.2024.grpo}. Crucially, to ground the generated reasoning in empirical accuracy, we utilize a deterministic reward function defined by the negative Mean Absolute Error (MAE) of the numerical forecast. This optimization objective ensures that the generated rationale serves as a functional intermediate representation aligned with accurate trajectory modeling, rather than merely plausible-sounding text (Appendix \ref{appendix:methods:reasoning}).

\subsection{Baselines}
We compare GDT against forecasting and event-prediction baselines (Appendix \ref{app:baselines}). Forecasting baselines include copy-forward, TiDE, and the foundation models Chronos and Chronos Bolt \cite{ansari.2024.chronos}, which were pretrained on over 700k time series. On the other hand, event prediction baselines include Random Survival Forest (RSF) and the EHR foundation model CLMBR-T, pretrained on 2.5 million patients and adapted with a Cox proportional hazards head. In real-world data experiments, baselines are trained on a representative 2k-patient subset; Appendix \ref{app:forecast_2k_section} shows that GDT retains its advantage under the same constraint. All clinical trial evaluations use identical datasets.

\section{Results}

We benchmark GDT across 20 cancer indications on real-world clinical data as well as on out-of-distribution clinical trial tasks, enabling evaluation in realistic settings.

\subsection{RWD Forecasting Results}

GDT consistently outperforms forecasting baselines across all indications. Aggregated across all variables, GDT achieves a median MASE of 0.867 (IQR 0.186), significantly outperforming the next-best model, multivariate TiDE (0.966; $p<0.001$, Wilcoxon signed-rank test). Other foundation models and baselines, including Chronos and Llama 3.1, exhibit higher error rates (Appendix \ref{app:forecasting_full}).

This performance advantage is even more pronounced on clinically critical, highly volatile variables. Evaluated on the top 30 most time-varying biomarkers per indication (Figure \ref{fig:cgdb:forecasting}), GDT achieves a median MASE of 0.830 (IQR 0.165), significantly outperforming multivariate TiDE (0.947; $p<0.001$). These results demonstrate GDT’s effectiveness in modeling dynamic patient trajectories relative to time-series models and general-purpose LLMs.

Notably, GDT retains this performance advantage even when trained on the same 2,000-patient subset as the baselines (Appendix \ref{app:forecast_2k_section}).

\subsection{RWD Event Results}

GDT shows statistically significant improvements across the majority of modeled clinical outcomes, specifically those with sufficient indication coverage (Figure \ref{fig:cgdb:events}). For survival, progression, and therapy switching (each with $N>10$ indications), GDT outperforms the strongest baseline, RSF, across all evaluated time points on Inverse Probability of Censoring Weighting (IPCW) C-index (0.703 vs 0.662), and is statistically significant in 11 of 12 time points ($p_{adj} < 0.05$). This advantage does not extend to time-to-metastasis (IPCW C-index 0.502 vs 0.563; $N=3$), where limited training diversity likely constrains performance, highlighting a low-data boundary for the proposed architecture.

Interestingly, we observe lower performance for the EHR foundation model CLMBR-T despite its larger pretraining cohort and task-specific fine-tuning. This may be attributable to its fixed vocabulary, which covers only 11 of the 283 therapies present in the dataset.


\begin{figure}
    \centering
    \includegraphics[width=0.8\linewidth]{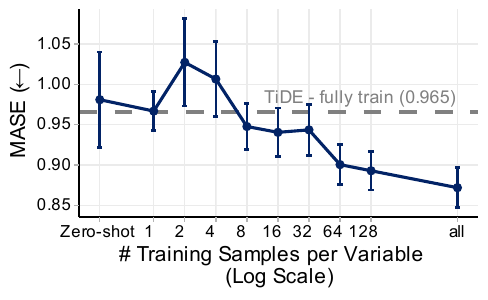}
    \caption{\textbf{Forecasting error as a function of training sample size.}  Evaluated on the test set of the POPLAR trial, using Mean Absolute Scaled Error (MASE), relative to training samples from the OAK trial (log X-axis; error bars denote IQR across 17 variables). Zero-shot performance is competitive, while low-data fine-tuning shows an initial error increase before stabilizing and improving at higher counts ($all \approx 880$/variable).}
    \label{fig:cit:samples_vs_performance}
\end{figure}

\subsection{Validation on Clinical Trials}

We further evaluate the method on out-of-distribution clinical trial tasks, using two trials for testing (POPLAR, IMpower130) and two additional trials for optional fine-tuning (OAK, IMpower131; Appendix \ref{app:setup:clinical_trials}). Clinical trials can be considered out-of-distribution due to differences in data measurement and format, as well as distinct patient populations, in comparison to the RWD training set \cite{orcutt.2025.rct_vs_rwd}. The task focuses on cold-start prediction from baseline measurements only, reflecting a challenging and clinically relevant trial setting. Note, the data is identical across all models in the same setup with respect to fine-tuning and zero-shot settings. GDT is evaluated in both zero-shot and fine-tuned settings, with identical data used across all models within each setup (Figure \ref{fig:cit:overall}).

\subsubsection{Clinical Trial Blood Biomarker Forecasting}

In the cold-start setting, GDT shows strong generalization performance. GDT, fine-tuned on OAK and IMpower131, then evaluated on POPLAR and IMpower130 attains the lowest error rates (median MASE 0.883 and 0.754), significantly outperforming the strongest baseline, univariate TiDE ($p<0.05$), while zero-shot GDT matches task-specific fine-tuned models (Appendix \ref{app:setup:clinical_trials}).

\subsubsection{Clinical Trial Event Prediction}

For clinical trial event prediction, fine-tuned GDT shows strong performance, achieving the highest C-index in all evaluation time points, with the zero-shot variant performing comparably high (IPCW C-indexes 0.672 and 0.656; Figure \ref{fig:cit:overall}). RSF provides a competitive baseline (IPCW C-index 0.648) without using longitudinal trajectories, while CLMBR-T shows variable performance across POPLAR and IMpower130, potentially reflecting vocabulary constraints. Statistical testing was not conducted due to limited sample sizes.

\begin{figure*}
    \centering
    \includegraphics[width=\linewidth]{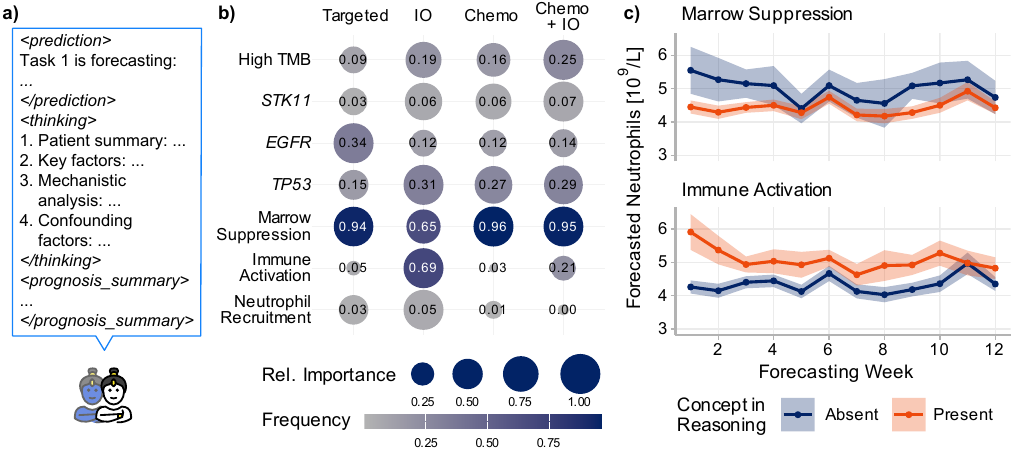}
    \caption{\textbf{GDT can be extended to provide clinically aligned reasoning.} \textbf{a)} The model generates a structured reasoning chain, including patient summary, key factors, and mechanistic analysis, alongside its numerical forecast. \textbf{b)} Keyword analysis shows alignment with clinical expectations, linking marrow suppression to chemotherapy, immune activation to immunotherapy (IO), and \textit{EGFR} mutations to targeted therapy \textbf{c)} Stratifying predicted neutrophil trajectories by presence of reasoning concepts confirms validated effects: ‘Marrow Suppression’ is linked to lower counts, while ‘Immune Activation’ is linked to higher counts.}
    \label{fig:reasoning:overall}
\end{figure*}

\subsubsection{Sample Efficiency Analysis}
We evaluate the sample efficiency of GDT in an out-of-distribution setting by training on the OAK trial and testing on POPLAR, which share identical treatment regimens (Figure \ref{fig:cit:samples_vs_performance}). Performance is assessed by progressively subsampling the training data across 17 target variables. 

While zero-shot performance is comparable to fully trained TiDE, fine-tuning GDT leads to consistently better performance at 64 samples per variable. Interestingly, the error increases initially, potentially hinting that training on few samples might overfit the model and be worse than the zero-shot setting.

\subsection{Reasoning Extension}

We demonstrate the flexibility of the TwinWeaver framework by extending GDT to generate interpretable clinical reasoning alongside forecasts for neutrophil trajectories in NSCLC (Figure \ref{fig:reasoning:overall}; Appendix \ref{app:reasoning:results}). 
Keyword analysis confirms that GDT generated reasoning chains correspond to clinically relevant features and genetic contexts, correctly associating high tumor mutational burden (TMB) more with immunotherapy based medications \cite{shao.2024.tmb}, while identifying \textit{EGFR} mutations as key drivers in targeted therapies and delineating distinct neutrophil trajectories based on patients' immune state \cite{dunbar.2021.genomic_thrombroembolic, orr.2007.bone_marrow}. The model additionally identifies \textit{STK11} as a key genetic mutation, which has been found to be associated with effects on neutrophil levels, as well as a prognostic marker \cite{zheng.2024.stk11_prognostic, koyama.2016.stk11_deficiency}.
Although the reasoning extension incurs a modest increase in forecasting error (MASE 0.862 vs 0.828), it offers a calculated trade-off for clinical utility by providing transparent, biologically grounded context for hypothesis generation.

\subsection{Training Synergy Across Cancer Indications}

We find that the pan-cancer training improves the performance of individual cancer indications, thus demonstrating the value of training across a diverse set of diseases, particularly for predictions in diseases that present less data (Appendix \ref{app:forecast_single_indication_overview}).

\subsection{Patient History Analysis}

We further show that incorporating richer patient histories can enhance clinical trial survival predictions beyond current data collection practices (Appendix \ref{app:ablation_input}). We select patients from the NSCLC test set with at least 52 weeks of history, and systematically cut patient history whilst comparing one-year survival IPCW C-index performance. We observe an improvement of 9.9\% of the C-index when using the full patient history, in comparison to using baseline information only, which is often done in the context of clinical trials.

\section{Conclusion}

Overall, our findings suggest that the TwinWeaver framework transforms pretrained LLMs using specific fine-tuning into clinical prediction engines, exemplified with GDT. Our benchmarks highlight that, despite standard methods remaining competitive baselines, fine-tuned LLMs overcome critical limitations regarding fixed vocabularies and multi-modal integration, thus improving prediction performance over existing methods. Specifically, the ability of GDT to generalize to new clinical trials as well as its pan-cancer capabilities offers a promising solution for low-data scenarios.

Future work should focus on systematic hyperparameter optimization, including sampling strategies and prompt design. Future iterations should also integrate modalities such as histopathology images, single-cell and spatial omics. Integrating reasoning directly into model training may enable reasoning-first clinical foundation models. From a translational perspective, rigorous clinical validation is required to assess real-world utility, and further refinement of time-to-event modeling is needed to produce well-calibrated survival curves while maintaining generative flexibility.

In conclusion, results with GDT demonstrate the effectiveness of the TwinWeaver framework, indicating that LLMs can be adapted successfully as a platform for patient digital twins.

\section*{Software and Data}

\alttext{\texttt{[Links to code redacted for anonymity.]}}{
All framework and model code can be found at \href{http://github.com/MendenLab/TwinWeaver}{http://github.com/MendenLab/TwinWeaver}.
}

The FH-FMI CGDB data that support the findings of this study were originated by and are the property of Flatiron Health, Inc. and Foundation Medicine, Inc. Requests for data sharing by license or by permission for the specific purpose of replicating results in this manuscript can be submitted to \alttext{\texttt{[emails redacted for anonymity]}}{\href{mailto:PublicationsDataAccess@flatiron.com}{PublicationsDataAccess@flatiron.com} and \href{mailto:cgdb-fmi@flatiron.com}{cgdb-fmi@flatiron.com}}. 

Qualified researchers may request access to individual clinical trial patient-level data for this analysis through a data request platform. At the time of writing, this request platform is \alttext{\texttt{[link redacted for anonymity.]}}{\href{https://vivli.org/ourmember/roche/}{https://vivli.org/ourmember/roche/}. For up to date details on Roche’s Global Policy on the Sharing of Clinical Information and how to request access to related clinical study documents, see here: \href{https://go.roche.com/data_sharing}{https://go.roche.com/data\_sharing}}. Anonymized records for individual patients across more than one data source external to \alttext{\texttt{[redacted]}}{Roche} cannot, and should not, be linked due to a potential increase in risk of patient re-identification. For the clinical trial cohort, 4 clinical studies sponsored by \alttext{\texttt{[redacted]}}{Roche/Genentech} were identified with the appropriate informed consent for secondary data use analysis. The analysis was conducted according to Good Clinical Practice and the Declaration of Helsinki.

\alttext{\section*{Acknowledgements} \texttt{Redacted for anonymous submission.}
}
{
\section*{Acknowledgements}
We thank the patients, families, caregivers, and site personnel who participated in the data from which were included in this analysis.

We also thank Pietro Belligoli, Clara Wong-Fannjiang, Maksim Eremeev, Jesse Swanson, Claudia Iriondo and Guy Amster for their feedback and valuable discussions. We would also like to thank Anna Bauer-Mehren, Monika Bug and Luise Wolf for supporting the collaborations, as well as Emilia Leogrande and Andrew Whitehead for enabling the publication.

This study was funded by F. Hoffmann-La Roche and the European Union's Horizon 2020 Research and Innovation Programme (Grant agreement No. 950293–COMBAT-RES). N.M., M.B. and M.P.M are supported by the Helmholtz Association under the joint research school “Munich School for Data Science - MUDS”.
}

\section*{Impact Statement}

This work advances precision oncology by enabling interpretable Patient Digital Twins via Large Language Models. By providing transparent reasoning alongside forecasts, TwinWeaver fosters clinical trust and offers a scalable solution for data-scarce scenarios in drug development. However, applying generative models to healthcare entails risks regarding hallucination and the propagation of biases inherent in Real-World Data. We emphasize that this framework is currently a research tool, rigorous prospective validation is an essential prerequisite before clinical deployment.

\onlyaccepted{
\section*{Consent for Publication}
This manuscript is sufficiently anonymised and does not contain any personal and/or medical information about an identifiable patient.
}

\onlyaccepted{
\section*{Competing Interests}
N.M., M.B., R.R.E., L.V.V. and F.S. are all employees of F. Hoffmann-La Roche. S.M., J.W., D.M. are employees of Genentech. M.P.M. collaborates and is financially supported by GSK, F. Hoffmann-La Roche, and AstraZeneca. M.P.M. consults Merck Sharp \& Dohme and McKinsey \& Company.
}

\onlyaccepted{\section*{Author Contributions}
N.M. conceived and executed the project. 
E.V.P., J.W., and D.M. contributed to methodology and the implementation of the software framework. 
M.B., L.V.V., S.M., and H.W. performed data preparation, experimental analysis and visualization.
K.~Choi, K.~Cho, and S.R. contributed to the methodology and study planning. 
R.R.E., F.S., and M.P.M. supervised, designed and directed the project.
All authors reviewed and contributed to the writing of the manuscript.
}

\bibliography{example_paper}
\bibliographystyle{icml2026}

\newpage
\appendix
\onecolumn



\section{Data Source}
\label{app:datssets:formal}

This study used the US-based deidentified Flatiron Health-Foundation Medicine Breast Cancer, Colorectal Cancer, Pancreatic Cancer, Ovarian Cancer, Metastatic Prostate Cancer, Gastric and Esophageal Cancer, Endometrial Cancer, Melanoma, Advanced Urothelial Carcinoma, Renal Cell Carcinoma, Head and Neck Cancer, Small Cell Lung Cancer, Multiple Myeloma, Hepatocellular Carcinoma, Acute Myeloid Leukemia, Chronic Lymphocytic Leukemia, Diffuse Large B-cell Lymphoma, and Follicular Lymphoma Clinico-Genomic Database (FH-FMI CGDB). 

Clinical data from the Flatiron Health Research Database \cite{Flatiron.2025.Guide} are linked to genomic data, derived from FMI’s comprehensive genomic profiling (CGP) tests (FoundationOne®CDx, FoundationOne®, FoundationOne®Liquid CDx, FoundationOne®Liquid and/or FoundationOne®Heme), in the FH-FMI CGDB by deterministic matching, providing a deidentified dataset \cite{Singal.2019.LungCancer}.

\newpage
\section{Methodological Details}
\label{app:methodology}

This section details the TwinWeaver framework, the specific GDT implementation details, and the inference post-processing methods.

\subsection{TwinWeaver Framework Details}

\subsubsection{Forecasting Variable Sampling}
\label{app:forecasting_sampling}

The probability $p(v)$ of sampling a variable $v \in V$ for the forecasting task is set proportionally to its count and its volatility. We define volatility using the root mean squared error (RMSE) of a simple ``copy-forward'' baseline, which predicts the next value will be the same as the last observed value. We filter variables to include only those which have at least a certain minimum number of observations, in the case of GDT, is set to 50.

Let $\mathcal{S}(v) = \bigcup_{p \in \mathcal{P}} \{ (\mathbf{m}_p(v, \tau_i), \mathbf{m}_p(v, \tau_{i+1})) \}$ be the set of all consecutive value pairs $(y, y')$ for $v$ (where $\tau_{i+1}$ is the next observation time after $\tau_i$ for patient $p$) across all patients. Let $N_v = |\mathcal{S}(v)|$. The copy-forward RMSE is:

$$ RMSE_v = \sqrt{ \frac{1}{N_v} \sum_{(y, y') \in \mathcal{S}(v)} (y' - y)^2 } $$

Let $\sigma_v$ be the standard deviation of all observed values of $v$. We define the normalized RMSE (NRMSE) as:

$$ NRMSE_v = \frac{RMSE_v}{\sigma_v} $$

Let $C_v$ be the total number of observations for variable $v$ in the dataset. The unnormalized sampling score $s_v$ is defined as:

$$ s_v = \log_2( C_v \times NRMSE_v ) $$

The final sampling probability $p(v)$ is obtained by normalizing these scores over all candidate variables in $V$. This strategy biases sampling towards variables that are frequently measured ($C_v$) and exhibit high, non-trivial variance over time ($NRMSE_v$), which are often the most clinically relevant and challenging to predict.

\subsubsection{Forecasting Details}
\label{app:forecasting_details}

The forecasting target $\mathcal{Y}^{\text{forecast}}$ is defined as the set of lists of future values for the sampled variables $V'$. We only include target values up to the forecast horizon and censor the prediction window at the start of any new competing event, such as a new line of therapy, which could fundamentally alter patient trajectory. Formally, the target is:
\begin{align*}
    \mathcal{Y}^{forecast}_p(t, V') =& \{ [\mathbf{m}_p(v, t+k) \mid 
    k \in \{1, \dots, \Delta t^{forecast}\},\\ 
    &t+k < \tau_{censor}(p, t)] \mid v \in V'\} 
\end{align*}

where $\mathbf{m}_p(v, \tau)$ is the value of lab $v$ at week $\tau$, $\tau_{censor}(p, t)$ is the time of the next competing event after $t$, and unmeasured values are treated as missing targets.

\paragraph{Additional Forecasting Task}
\label{app:forecasting_bucket}
To ensure the model remains relevant across a multitude of tasks, we also include a forecasting task for binned values. In this setup, the task is to predict which of 5 equally numbered quintiles the variable's value lands in across the future time steps. However, due to its relevance and scope, we do not further evaluate this task in this paper, and it should be explored in future work.

\subsubsection{Landmark Events Details}
\label{app:landmark_events_details}

\begin{algorithm}[h!]
\caption{TwinWeaver: Landmark Event Sampling Strategy}
\label{alg:landmark-sampling}
\begin{algorithmic}[1]
\REQUIRE Patient $p$, history $\mathcal{X}_p(t)$ up to split time $t$, event set $\mathcal{E}$, horizon range $h$
\ENSURE Training tuple $(\mathcal{X}_p(t), E, \Delta t^{event}, y)$ with $y \in Y$
\STATE Randomly sample an event $E \in \mathcal{E}$ (e.g., disease progression).
\STATE Sample a horizon $\Delta t^{event} \sim \mathrm{Uniform}[1,h]$.
\STATE Set $t^{event} \leftarrow t + \Delta t^{event}$.
\STATE Determine whether $t^{event}$ is censored (e.g., by new therapy) and whether $E$ occurred by $t^{event}$:
\begin{itemize}
    \item If a censoring event occurs in $(t, t^{event}]$ before occurrence of event $E$, set $y \leftarrow \text{censored}$.
    \item Else if $E$ occurs by $t^{event}$, set $y \leftarrow \text{occurred}$.
    \item Else set $y \leftarrow \text{not occurred}$.
\end{itemize}
\STATE \textbf{return} $(\mathcal{X}_p(t), E, \Delta t^{event}, y)$.
\end{algorithmic}
\end{algorithm}

The uniform sampling of the horizon $\Delta t$ acts as a form of data augmentation, training the model to generalize to any on-demand prognostic query at inference time. Although computationally feasible, this sampling approach prevents overfitting to canonical time points (e.g., 1-year survival) and encourages a more robust understanding of time-aware risk.

We also recognize that a uniform sampling strategy could be questioned, as clinical event hazards are rarely uniform. This mismatch could introduce a bias, potentially leading the model to be poorly calibrated at extreme horizons. Given the large capacity of the LLM and the millions of training samples, we hypothesize that this bias is minimized, but we explicitly evaluate this by testing on stratified, predetermined landmark times.

\subsubsection{Combining Prompt}
\label{app:combining_prompt}

\begin{algorithm}[h!t]
\caption{TwinWeaver: Prompt Construction for Supervised Fine-Tuning}
\label{alg:prompt-construction}
\begin{algorithmic}[1]
\REQUIRE Split date $t$, patient history $\mathcal{X}_p(t)$ up to split time $t$, forecasting variables $V' \subseteq V$, targets $\mathcal{Y}^{forecast}_p(t)$ containing forecasting targets $\mathcal{Y}_p^{forecast}(t^{forecast}, V')$, at least one event setup $\{(E_j,\Delta t_j,y_j)\}_j$.
\ENSURE Input string $S_{\text{prompt}}$ and target completion $S_{\text{target}}$
\STATE Encode a fixed starting prompt $\pi_0$ into $S_{\text{prompt}}$.
\STATE Encode the patient's constant data ($\mathbf{s}_p$; e.g., demographics).
\STATE Encode the history $\mathcal{H}_p(t)$ visit-by-visit in chronological order; alphabetically sort items within each visit. All missing values are skipped. If total size of prompt exceeds context length, we omit visits until the total prompt is less than the context length, while ensuring to include the first visit, which often contains important initial diagnoses.
\STATE Repeat the latest critical observations (e.g. genetic observations, the latest line of therapy, and the last observed values of all $v \in V'$), to avoid the lost in the middle problem.
\STATE Encode all tasks in text and assign each a numeric task ID $j=1,\dots,|\mathcal{T}|$.
\STATE Construct $S_{\text{target}}$ by encoding the corresponding task answers in the same order and required format, skipping missing data.
\STATE \textbf{return} $(S_{\text{prompt}}, S_{\text{target}})$.
\end{algorithmic}
\end{algorithm}

In Algorithm \ref{alg:prompt-construction}, we see how the final full prompt is constructed. Note, that whilst the approach is limited by context length, we anticipate that this will become less of an issue as the current large, state of the art models can process over 100k tokens.


\subsection{Inference Post-Processing}

\subsubsection{Constructing Monotonic Cumulative Incidence Function (Isotonic Regression)}
\label{app:isotonic_regression}

Whereas the landmarking approach described in Section \ref{twin_weaver:events_inference} provides robust risk estimates for specific horizons, treating each horizon $\Delta t$ as an independent query can result in temporal inconsistencies. To recover a consistent Cumulative Incidence Function (CIF) $F(t)$ for individual patient analysis, we employ a two-step post-processing strategy: \textit{conditioning} followed by \textit{isotonic calibration}.

\textbf{Step 1: Conditioning.} We first isolate the risk signal from the censoring distribution. Under the assumption of non-informative censoring, the probability of the event occurring by time $t$, given that the observation was not censored before $t$, is a consistent estimator of the true cumulative incidence. We compute this conditional probability $\hat{y}_i$ for each horizon $\tau_i$ by renormalizing the model's softmax outputs over the known outcomes:
$$ \hat{y}_i = \frac{P(\text{occurred} \mid \tau_i)}{P(\text{occurred} \mid \tau_i) + P(\text{not occurred} \mid \tau_i)} $$

\textbf{Step 2: Isotonic Calibration.} As the independent estimates $\hat{y}_i$ are not guaranteed to be monotonic, we solve for the calibrated sequence $y^*_1, \dots, y^*_K$ for a set of $K$ landmark times that minimizes the squared error subject to the monotonicity constraint (Cumulative Incidence cannot decrease):
\begin{align*}
\min_{y^*} \sum_{i=1}^K (y^*_i - \hat{y}_i)^2 \quad \text{subject to} \quad y^*_i \le y^*_{i+1} \quad \forall i
\end{align*}

We implement this using the Pool Adjacent Violators Algorithm (PAVA). This effectively projects the noisy LLM predictions onto the space of valid probability distribution functions, ensuring clinical interpretability without retraining the model.

\newpage
\section{Experimental Setup}
\label{app:setup}

\subsection{Dataset Statistics}

\label{app:data:statistics}

\begin{table}[htbp]
    \centering
    \caption{Dataset Statistics: Patient Counts, Biomarkers, and Prediction Capabilities by Indication}
    \label{tab:dataset_stats}
    \scriptsize 
    \begin{tabular}{llrrcccc}
        \toprule
         & & & & \multicolumn{4}{c}{\textbf{Event Types Predicted}} \\
        \cmidrule(lr){5-8}
        \textbf{Indication} & \textbf{Abbr.} & \textbf{Patients} & \textbf{For. Biomarkers} & \textbf{Surv.} & \textbf{Prog.} & \textbf{Switch} & \textbf{Met.} \\
        \midrule
        Undetermined/multiple & Und. & 19,339 & 90 & \checkmark & & \checkmark & \\
        Non-Small Cell Lung Cancer & NSCLC & 17,834 & 74 & \checkmark & \checkmark & \checkmark & \checkmark \\
        Breast Cancer & BC & 13,901 & 87 & \checkmark & \checkmark & \checkmark & \checkmark \\
        Colorectal Cancer & CRC & 11,085 & 80 & \checkmark & \checkmark & \checkmark & \\
        Ovarian Cancer & OC & 5,423 & 82 & \checkmark & \checkmark & \checkmark & \\
        Metastatic Prostate Cancer & mPC & 5,251 & 78 & \checkmark & & \checkmark & \\
        Pancreatic Cancer & Panc & 4,506 & 69 & \checkmark & & \checkmark & \\
        Gastric / Esophageal Cancer & GE & 3,826 & 67 & \checkmark & \checkmark & \checkmark & \\
        Endometrial Cancer & EC & 2,335 & 78 & \checkmark & & \checkmark & \\
        Advanced Urothelial Carcinoma & aUC & 2,071 & 62 & \checkmark & \checkmark & \checkmark & \\
        Melanoma & MEL & 1,951 & 72 & \checkmark & \checkmark & \checkmark & \checkmark \\
        Renal Cell Carcinoma (Kidney) & RCC & 1,385 & 67 & \checkmark & \checkmark & \checkmark & \\
        Head and Neck Squamous Cell & HNSCC & 934 & 55 & \checkmark & & \checkmark & \\
        Small Cell Lung Cancer & SCLC & 852 & 53 & \checkmark & \checkmark & \checkmark & \\
        Multiple Myeloma & MM & 564 & 87 & \checkmark & \checkmark & \checkmark & \\
        Acute Myeloid Leukemia & AML & 504 & 67 & \checkmark & & \checkmark & \\
        Hepatocellular Carcinoma (Liver) & HCC & 405 & 50 & \checkmark & \checkmark & \checkmark & \\
        Diffuse Large B-Cell Lymphoma & DLBCL & 350 & 57 & \checkmark & & \checkmark & \\
        Chronic Lymphocytic Leukemia & CLL & 337 & 55 & \checkmark & & \checkmark & \\
        Follicular Lymphoma & FL & 201 & 50 & \checkmark & & \checkmark & \\
        \bottomrule
    \end{tabular}
    
    \vspace{1ex}
    {\raggedright \textit{\footnotesize Note: For. = Forecasted, Surv. = Survival, Prog. = Progression, Switch = Switching Treatments, Met. = Metastasis.} \par}
\end{table}
\normalsize

\begin{table}[ht]
    \centering
    \caption{Patient demographics and visit statistics by indication. Drug visits is the number of visits where a drug was administered for a given patient.}
    \label{tab:indication_stats}
    \begin{tabular}{lrrrr}
        \toprule
        \textbf{Indication} & \textbf{Birth Year} & \textbf{Gender} & \textbf{Visits} & \textbf{Drug Visits} \\
         & (mean) & (\% Female) & (mean) & (mean) \\
        \midrule
        AML & 1954.1 & 42.9 & 132.6 & 37.1 \\
        aUC & 1949.2 & 28.8 & 67.8 & 21.2 \\
        BC & 1958.5 & 98.9 & 94.7 & 32.6 \\
        CLL & 1951.9 & 38.0 & 116.9 & 18.4 \\
        CRC & 1957.8 & 45.7 & 92.8 & 29.1 \\
        DLBCL & 1955.4 & 39.7 & 89.0 & 17.3 \\
        EC & 1952.7 & 100.0 & 79.6 & 20.6 \\
        FL & 1955.4 & 45.3 & 98.6 & 24.7 \\
        GE & 1954.3 & 26.7 & 66.7 & 20.1 \\
        HCC & 1953.4 & 24.9 & 54.2 & 9.9 \\
        HNSCC & 1955.3 & 21.0 & 83.1 & 24.0 \\
        MEL & 1954.7 & 34.6 & 61.2 & 17.5 \\
        mPC & 1948.6 & 0.0 & 65.2 & 12.0 \\
        MM & 1953.3 & 42.6 & 145.9 & 66.9 \\
        NSCLC & 1950.8 & 51.1 & 62.3 & 18.1 \\
        OC & 1954.3 & 100.0 & 100.3 & 29.1 \\
        Panc & 1953.0 & 45.5 & 68.7 & 18.6 \\
        Undetermined & 1956.7 & 50.1 & 73.3 & 19.2 \\
        RCC & 1955.9 & 28.5 & 73.2 & 20.1 \\
        SCLC & 1952.5 & 51.2 & 68.3 & 27.2 \\
        \bottomrule
    \end{tabular}
\end{table}
\normalsize

\subsubsection{Data Preprocessing Details}
\label{app:data_preprocessing}

Our preprocessing pipeline converts raw multi-modal data such as diagnoses, genetic tests, labs, etc., into a weekly longitudinal format. Importantly, we encode the patient's genetic information as text as well, noting somatic variants identified from targeted sequencing, including details of the specific mutation.

We include tables on diagnoses, drug administrations, ECOG scores, progression, mortality, response, metastases, surgery, labs, vitals, demographics, cancer-specific biomarkers, extracted genetic signatures, basic biomarkers, and paneled genetic data. While the raw datasets are similar, each required specific adjustments for tables and columns. We aggregate all events on a weekly level to align with the typical frequency of cancer therapy events and avoid over-representing acute hospital visits. Numeric and categorical variables are aggregated using their mean and mode, respectively.

To reduce the issue of outliers in the forecasting task, we apply 3-sigma filtering to all target blood lab values during preprocessing \cite{pukelsheim.1994.three_sigma}. When encoding numerical values into strings, we limit the number of post-decimal digits to 2 to prevent single long numbers from dominating the string. Note that a small number of patients (~311) that were present in multiple indication datasets were discarded from the test and validation sets.

\normalsize
\subsection{GDT Model \& Training Details}

\subsubsection{Training Details}
\label{app:training_details}

We limit the total context length to 8000 tokens to manage GPU memory constraints, truncating the earliest visits if necessary while always preserving the first.

For training, we generate $L$ number of splits per patient per line of therapy. We experimented with $L \in \{1, 10\}$ and observed a lower validation loss with $L=10$. This higher sampling rate per patient increases the total sample count to 2.49 million and provides the model with more diverse examples from each patient's trajectory. In future work with more computational resources, higher sampling rates can also be explored.  For outlier filtering, we also apply 3-sigma filtering, based on the per-variable statistics using the training dataset, following \cite{makarov.2025.llms}.

For all smaller fine-tuning runs, we also explore using the cosine learning rate scheduler as well as a 10\% warm up ratio, determined then based on the validation loss of the respective run.

For the fine-tuning, formally, let the full text sequence be $x = (x_1, \dots, x_N)$, where $x_{\text{prompt}} = (x_1, \dots, x_{k-1})$ and $x_{\text{target}} = (x_k, \dots, x_N)$. The loss function is:
$$ \mathcal{L}(\theta) = - \sum_{i=k}^{N} \log q_\theta(x_i \mid x_{1:i-1}) $$
where $q_\theta$ is the LLM parameterized by $\theta$. 

The core training run was performed across 8 H100 GPUs, taking around 7 days, corresponding to 178 GPU hours, with full fine-tuning. The per-device batch size was 1, resulting in an effective batch size of 8. 

\subsubsection{Reasoning Extension Method}
\label{appendix:methods:reasoning}

Using fine-tuned LLMs as a prediction platform enables a variety of downstream tasks, including further fine-tuning to enable reasoning. Understanding the reasons behind a certain prediction is a critical task in biomedicine. In our example, we focus on the largest single indication NSCLC. We chose to focus on forecasting the development of neutrophil counts in the peripheral blood owing to their high measurement frequency and variability. We demonstrate this capability by extracting 3000 random patients from the NSCLC train dataset. We also extract 300 validation and test set patients. For all patients we then generate the training data as outlined above. After dropping all patients without enough neutrophil observations, we obtained 2385 train, 235 validation and 244 test set patients.

For the reasoning, we use a setting, in which we first generate the prediction and then provide the potential reasoning. As in most cases the literature to generate a truly logical step-by-step approach is limiting, we consider this step a hypothesis-generating approach. Additionally, in this setting, it is possible to provide a specific prediction and generate multiple possible explanations, an important tool to provide different hypotheses for a patient's response in clinical practice.

We first generated a reasoning dataset by prompting the Qwen3 Next 80B-A3B model to complete a reasoning chain given a patient's history and the true targets. We then applied deterministic basic filtering (e.g. formatting). 

We then trained GDT on this dataset for the initial predictions. Finally, to improve reasoning and prediction performance, we then applied GRPO \cite{shao.2024.grpo} to the fine-tuned reasoning model. The GRPO reward function used only the mean absolute error, with a lower value leading to a higher reward, and capped for errors above 20 units. $\beta = 0$, the learning rate was set to $10^{-6}$, using 8 generations, 0.1 warm up ratio, 1 epoch.

\subsection{Baselines \& Metrics}
\label{app:baselines}

\subsubsection{Forecasting Baselines}
\label{app:baselines:forecasting}

We compare GDT with standard time-series models on the blood biomarker forecasting task. 

To ensure fair comparisons, we performed extensive post-processing of the data for the baseline models. For the missingness, we performed linear interpolation together with forward and backward filling for the input time series, and forward filling for the target variable. Since potentially other lab variables are relevant for the prediction, we include other historical time series of all lab variables overlapping across all cancer types. This results in 74 lab variables for the real-world pan-cancer dataset and 43 variables in the clinical trials dataset. We use these variables as inputs in the multivariate capable models, making sure there is no temporal data leakage. We additionally provide a static feature vector with additional variables, using their last observed value if needed, including age, gender, number of diagnoses, number of genetic events, therapy name, therapy number and key blood biomarkers. We apply a 30-week look-back window, and 13-week prediction window, ensuring that the model sees over half a year of patient data. More historic blood lab value information further than 30 weeks into the past is most likely not producing additional information relevant for blood biomarker prediction. As in GDT, we also apply 3-sigma filtering to the data to remove outliers. 

Evaluating baselines on the full pan-cancer dataset was constrained by the heterogeneity of the data structure (variable lengths, missingness patterns) which required extensive preprocessing for non-LLM architectures. We therefore utilized a random subsample, 2000 training patients for the baselines, resulting in 45,023 training samples with multiple lines of therapies and variables per sample. Note, in the clinical trial tasks, all fine-tuned models used identical training datasets.

To ensure a fair comparison, we then used an identical number of training patients and set up on the clinical trial validation data sets as for the GDT model, see Appendix \ref{app:forecast_2k_section} for details. However, GDT still has improved performance.

TiDE, Chronos and Chronos Bolt are all implemented in AutoGluon \cite{autogluon}. For TiDE, we explored both, including and excluding all helper columns and a hyperparameter search for the learning rate. TiDE also processes the static features. Chronos and Chronos Bolt use only the target variable's history as input, since the models cannot process multiple input time series and we observed a lower validation loss when excluding the static features vector. We believe this is due to the implementation in AutoGluon, which requires an external regressor. For Chronos fine-tuning, we used the default learning rate of $10^{-3}$ and performed early stopping.

For the clinical trial cold start task, where only baseline measurements were available, we constructed a synthetic history for the time-series baselines (TiDE, Chronos) to satisfy their input window requirements. We backfilled the baseline value across the 30-week look-back window. This provided the models with a valid input tensor representing a constant history. This setup specifically evaluates the models' ability to infer future trajectory dynamics based on static covariates and baseline magnitude, rather than historical volatility.

\paragraph{Metrics}

We evaluate forecasting performance primarily using the Mean Absolute Scaled Error (MASE), with the copy forward as the naive predictor. MASE is particularly suitable for multi-modal clinical datasets where biomarker scales differ by orders of magnitude (e.g., neutrophils vs. albumin). We compute the metric by first aggregating the error across all patient samples for a given variable and horizon, and then taking the ratio against the aggregated copy forward (naive) baseline error. A MASE $< 1$ indicates that the model provides superior forecasts to simply carrying forward the last known value. We formally define the aggregated MASE for a specific variable $v$ as:

$$\mathrm{MASE}_v = \frac{\sum_{p \in \mathcal{P}} \sum_{j=1}^J |y_{p,j} - \hat{y}_{p,j}|}{\sum_{p \in \mathcal{P}} \sum_{j=1}^J |y_{p,j} - y_{p,\text{last}}|}$$

where $p$ indexes the patients, $J$ is the number of forecasting horizons (up to 13 weeks in this case), $y_{p,j}$ is the true value, $\hat{y}_{p,j}$ is the predicted value, and $y_{p,\text{last}}$ is the last observed value before the split time $t$. We report the median of these $\mathrm{MASE}_v$ scores across all variables and indications to ensure robustness against outliers common in real-world EHR data. We apply 3-sigma capping to minimize the effect of outliers.

Note, we evaluate forecasting performance using a population-aggregated variant of the standard Mean Absolute Scaled Error (MASE). Standard MASE \cite{hyndman.2006.mase} typically scales the error of each individual time series by its own in-sample historical volatility. However, in the context of sparse and irregular clinical data, patient histories are often short or exhibit periods of stability, leading to near-zero denominators and numerical instability when calculating per-patient ratios.

Since time-varying variables are the most relevant in clinical settings, we define the 30 most time varying variables, based on the mean absolute percentage error of the copy-forward baseline.

\subsubsection{Events Baselines}
\label{app:baselines:events}

We use the Random Survival Forest as a strong standard machine learning survival baseline focusing on the last observed values. For Random Survival Forest, we take the last observed values of critical patient parameters, representing a standard survival model. Specifically, it includes the following parameters as input: Indication, Gender, Events Until Split, Age, Number of Previous Diagnoses, Number of Genetic Events, Values from ROPRO Constants \cite{becker.2020.ropro}, Therapy Name, Therapy Line Number, Weight, Height, Oxygen Saturation, Systolic Blood Pressure, ECOG, Hemoglobin, Urea Nitrogen, Platelets, Calcium, Glucose, Lymphocytes Percent, Alkaline Phosphatase, Total Protein, Alanine Aminotransferase, Albumin, Total Bilirubin, Chloride, Monocytes Number, Eosinophils Percent, and Lactate Dehydrogenase. We use the `scikit-survival` library, encoding all categorical variables using a one hot encoder and using default hyperparameters.

To compare with state of the art EHR foundation models, we employed CLMBR-T. Using CLMBR-T required extensive preprocessing of the data, as the vocabulary of the model is fixed. We transformed all diagnoses, medications, LOINC and demographic data into the MEDS-OMOP format. We then used CLMBR-T through the femr package \cite{femr.2024} to generate the patient embeddings, and then trained a Cox’s proportional hazard’s model with elastic net penalty from scikit-survival. We applied a grid search with cross validation to find the optimal hyperparameters. 

Following the forecasting baselines, we use a 2000 patient subset for training on the pan-cancer dataset, and the full training set for the clinical trials.

\paragraph{Metrics} 
For the landmark event tasks, we assess the model's ability to correctly rank patients by risk using the Inverse Probability of Censoring Weighting (IPCW) Concordance Index (C-index)\cite{uno.2011.ipcw}. The standard C-index is biased in the presence of right-censored data; IPCW corrects for this by weighting patient pairs by the inverse probability of remaining in the study, estimated via the Kaplan-Meier estimator of the censoring distribution. Formally, for a pair of patients $i$ and $j$, it measures the probability that the patient with the higher predicted risk score experiences the event first:

$$\text{C-index} = \frac{\sum_{i,j} \mathbb{I}(T_i < T_j) \cdot \mathbb{I}(\hat{R}_i > \hat{R}_j) \cdot W_{ij}}{\sum_{i,j} \mathbb{I}(T_i < T_j) \cdot W_{ij}}$$

where $T$ is the event time, $\hat{R}$ is the predicted risk score (probability of event occurrence), and $W_{ij}$ are the censoring weights. A C-index of $0.5$ implies random ranking, while $1.0$ implies perfect discrimination. To further evaluate the reliability of the model's confidence, we additionally compute the Brier Score, which measures the mean squared difference between the predicted probabilities and the actual outcomes \cite{graf.1999.brier}. Lower Brier scores indicate better calibration. The Brier Score results on the adjusted monotonic versions are detailed in Appendix \ref{app:isotonic_regression}.

\subsection{Evaluation Inference Details}
\label{app:inference_details}

For the forecasting inference on the pan-cancer test set, we set $L=3$, ensuring that over 90\% of variables are evaluated in every indication, whilst still keeping the computational costs low.

Note, we observed that GDT outputs consistently, achieving a 0\% formatting and syntax errors across all indications in the real-world dataset.

\subsubsection{Evaluated Variables (LOINC Codes)}

\small
\begin{longtable}{lcccccccccccccccccccc}
\caption{Evaluated forecasting variables by LOINC code across indications.} \\
\toprule
LOINC & \rotatebox{90}{AML} & \rotatebox{90}{BC} & \rotatebox{90}{CLL} & \rotatebox{90}{CRC} & \rotatebox{90}{DLBCL} & \rotatebox{90}{EC} & \rotatebox{90}{FL} & \rotatebox{90}{GE} & \rotatebox{90}{HCC} & \rotatebox{90}{HNSCC} & \rotatebox{90}{MEL} & \rotatebox{90}{MM} & \rotatebox{90}{NSCLC} & \rotatebox{90}{OC} & \rotatebox{90}{Und.} & \rotatebox{90}{RCC} & \rotatebox{90}{SCLC} & \rotatebox{90}{Undet.} & \rotatebox{90}{aUC} & \rotatebox{90}{mPC} \\
\midrule
\endfirsthead

\multicolumn{21}{c}%
{{\bfseries \tablename\ \thetable{} -- continued from previous page}} \\
\toprule
Variable & \rotatebox{90}{AML} & \rotatebox{90}{BC} & \rotatebox{90}{CLL} & \rotatebox{90}{CRC} & \rotatebox{90}{DLBCL} & \rotatebox{90}{EC} & \rotatebox{90}{FL} & \rotatebox{90}{GE} & \rotatebox{90}{HCC} & \rotatebox{90}{HNSCC} & \rotatebox{90}{MEL} & \rotatebox{90}{MM} & \rotatebox{90}{NSCLC} & \rotatebox{90}{OC} & \rotatebox{90}{Panc} & \rotatebox{90}{RCC} & \rotatebox{90}{SCLC} & \rotatebox{90}{Undet.} & \rotatebox{90}{aUC} & \rotatebox{90}{mPC} \\
\midrule
\endhead

\midrule
\multicolumn{21}{r}{{Continued on next page}} \\
\bottomrule
\endfoot

\bottomrule
\endlastfoot

10334-1 &  &  &  &  &  &  &  &  &  &  &  &  &  & \cmark &  &  &  & \cmark &  &  \\
11051-0 &  &  &  &  &  &  &  &  &  &  &  & \cmark &  &  &  &  &  &  &  &  \\
13992-3 &  &  &  &  &  &  &  &  &  &  &  & \cmark &  &  &  &  &  &  &  &  \\
14804-9 &  &  &  &  & \cmark &  &  & \cmark &  &  & \cmark & \cmark &  & \cmark &  & \cmark &  & \cmark &  &  \\
14979-9 & \cmark & \cmark &  & \cmark & \cmark & \cmark & \cmark & \cmark &  & \cmark & \cmark & \cmark & \cmark & \cmark & \cmark & \cmark &  & \cmark & \cmark & \cmark \\
15152-2 &  &  &  &  &  & \cmark &  &  &  &  &  & \cmark & \cmark &  &  &  &  & \cmark &  &  \\
1742-6 & \cmark & \cmark & \cmark & \cmark & \cmark & \cmark & \cmark & \cmark & \cmark & \cmark & \cmark & \cmark & \cmark & \cmark & \cmark & \cmark & \cmark & \cmark & \cmark & \cmark \\
1751-7 & \cmark & \cmark & \cmark & \cmark & \cmark & \cmark & \cmark & \cmark & \cmark & \cmark & \cmark & \cmark & \cmark & \cmark & \cmark & \cmark & \cmark & \cmark & \cmark & \cmark \\
17817-8 &  &  &  &  &  &  &  &  &  &  &  & \cmark &  &  &  &  &  &  &  &  \\
17842-6 &  & \cmark &  &  &  &  &  &  &  &  &  &  &  &  &  &  &  & \cmark &  &  \\
17861-6 & \cmark & \cmark & \cmark & \cmark & \cmark & \cmark & \cmark & \cmark & \cmark & \cmark & \cmark & \cmark & \cmark & \cmark & \cmark & \cmark & \cmark & \cmark & \cmark & \cmark \\
18262-6 &  & \cmark &  & \cmark &  & \cmark &  &  &  &  & \cmark &  &  &  & \cmark & \cmark &  &  &  &  \\
1834-1 &  &  &  &  &  &  &  &  & \cmark &  &  &  &  &  &  &  &  &  &  &  \\
19023-1 & \cmark & \cmark & \cmark & \cmark &  & \cmark & \cmark & \cmark &  &  & \cmark & \cmark & \cmark & \cmark & \cmark & \cmark & \cmark & \cmark & \cmark & \cmark \\
19123-9 & \cmark & \cmark & \cmark & \cmark & \cmark & \cmark & \cmark & \cmark & \cmark & \cmark & \cmark & \cmark & \cmark & \cmark & \cmark & \cmark & \cmark & \cmark & \cmark & \cmark \\
1920-8 & \cmark & \cmark & \cmark & \cmark & \cmark & \cmark & \cmark & \cmark & \cmark & \cmark & \cmark & \cmark & \cmark & \cmark & \cmark & \cmark & \cmark & \cmark & \cmark & \cmark \\
1952-1 &  &  &  &  &  &  &  &  &  &  &  & \cmark &  &  &  &  &  & \cmark &  &  \\
1963-8 &  &  &  & \cmark &  & \cmark &  &  &  &  &  &  &  & \cmark &  &  &  & \cmark &  &  \\
1968-7 &  & \cmark &  & \cmark &  & \cmark &  & \cmark & \cmark & \cmark &  & \cmark & \cmark & \cmark & \cmark & \cmark &  & \cmark & \cmark & \cmark \\
1971-1 &  & \cmark &  &  & \cmark & \cmark &  &  &  & \cmark & \cmark & \cmark & \cmark & \cmark &  & \cmark &  &  &  & \cmark \\
1975-2 & \cmark & \cmark & \cmark & \cmark & \cmark & \cmark & \cmark & \cmark & \cmark & \cmark & \cmark & \cmark & \cmark & \cmark & \cmark & \cmark & \cmark & \cmark & \cmark & \cmark \\
1988-5 &  &  &  & \cmark &  &  &  &  &  &  & \cmark & \cmark &  &  & \cmark &  &  & \cmark &  &  \\
1994-3 &  &  &  &  &  &  &  &  &  &  &  &  &  &  &  &  &  & \cmark &  &  \\
2028-9 & \cmark & \cmark & \cmark & \cmark & \cmark & \cmark & \cmark & \cmark & \cmark & \cmark & \cmark & \cmark & \cmark & \cmark & \cmark & \cmark & \cmark & \cmark & \cmark & \cmark \\
2039-6 &  & \cmark &  & \cmark &  &  &  & \cmark &  &  &  &  & \cmark & \cmark & \cmark &  &  & \cmark &  & \cmark \\
20482-6 & \cmark & \cmark & \cmark & \cmark &  & \cmark & \cmark &  &  &  & \cmark &  & \cmark & \cmark & \cmark & \cmark &  & \cmark & \cmark & \cmark \\
20570-8 & \cmark & \cmark & \cmark & \cmark & \cmark & \cmark & \cmark & \cmark & \cmark & \cmark & \cmark & \cmark & \cmark & \cmark & \cmark & \cmark & \cmark & \cmark & \cmark & \cmark \\
2075-0 & \cmark & \cmark & \cmark & \cmark & \cmark & \cmark & \cmark & \cmark & \cmark & \cmark & \cmark & \cmark & \cmark & \cmark & \cmark & \cmark & \cmark & \cmark & \cmark & \cmark \\
2085-9 &  & \cmark &  &  &  & \cmark &  &  &  &  &  &  & \cmark & \cmark &  &  &  & \cmark & \cmark & \cmark \\
2093-3 &  & \cmark &  & \cmark &  &  &  &  &  &  &  &  & \cmark & \cmark & \cmark &  &  & \cmark & \cmark & \cmark \\
2141-0 &  &  &  &  &  &  &  &  &  &  & \cmark &  &  &  &  &  &  &  &  &  \\
2143-6 &  &  &  &  &  &  &  &  &  &  & \cmark &  &  &  &  &  &  &  &  &  \\
2160-0 & \cmark & \cmark & \cmark & \cmark & \cmark & \cmark & \cmark & \cmark & \cmark & \cmark & \cmark & \cmark & \cmark & \cmark & \cmark & \cmark & \cmark & \cmark & \cmark & \cmark \\
2161-8 &  &  &  &  &  &  &  & \cmark &  &  &  & \cmark &  &  &  & \cmark &  &  &  & \cmark \\
2162-6 &  &  &  &  &  &  &  &  &  &  &  & \cmark &  &  &  &  &  &  &  &  \\
2276-4 & \cmark & \cmark & \cmark & \cmark & \cmark & \cmark &  & \cmark &  & \cmark & \cmark & \cmark & \cmark & \cmark & \cmark & \cmark &  & \cmark & \cmark & \cmark \\
2324-2 &  & \cmark &  &  &  &  &  & \cmark &  &  &  & \cmark & \cmark & \cmark &  &  &  &  &  &  \\
2345-7 & \cmark & \cmark & \cmark & \cmark & \cmark & \cmark & \cmark & \cmark & \cmark & \cmark & \cmark & \cmark & \cmark & \cmark & \cmark & \cmark & \cmark & \cmark & \cmark & \cmark \\
24108-3 &  & \cmark &  & \cmark &  &  &  & \cmark &  &  &  &  & \cmark &  & \cmark &  &  & \cmark &  &  \\
2458-8 & \cmark &  & \cmark &  & \cmark &  &  &  &  &  &  & \cmark &  &  &  &  &  & \cmark &  &  \\
2465-3 &  & \cmark &  & \cmark & \cmark &  &  &  &  &  &  & \cmark &  &  &  &  &  &  &  &  \\
2472-9 &  &  & \cmark &  &  &  &  &  &  &  &  & \cmark &  &  &  &  &  & \cmark &  &  \\
2532-0 & \cmark & \cmark & \cmark & \cmark & \cmark & \cmark & \cmark & \cmark &  & \cmark & \cmark & \cmark & \cmark & \cmark & \cmark & \cmark & \cmark & \cmark & \cmark & \cmark \\
2571-8 &  & \cmark &  & \cmark &  & \cmark &  &  &  &  &  &  &  &  &  & \cmark &  & \cmark & \cmark & \cmark \\
26444-0 & \cmark & \cmark & \cmark & \cmark & \cmark & \cmark & \cmark & \cmark & \cmark & \cmark & \cmark & \cmark & \cmark & \cmark & \cmark & \cmark & \cmark & \cmark & \cmark & \cmark \\
26449-9 & \cmark & \cmark & \cmark & \cmark & \cmark & \cmark & \cmark & \cmark & \cmark & \cmark & \cmark & \cmark & \cmark & \cmark & \cmark & \cmark & \cmark & \cmark & \cmark & \cmark \\
26450-7 & \cmark & \cmark & \cmark & \cmark & \cmark & \cmark & \cmark & \cmark & \cmark & \cmark & \cmark & \cmark & \cmark & \cmark & \cmark & \cmark & \cmark & \cmark & \cmark & \cmark \\
26453-1 & \cmark & \cmark & \cmark & \cmark & \cmark & \cmark & \cmark & \cmark & \cmark & \cmark & \cmark & \cmark & \cmark & \cmark & \cmark & \cmark & \cmark & \cmark & \cmark & \cmark \\
26464-8 & \cmark & \cmark & \cmark & \cmark & \cmark & \cmark & \cmark & \cmark & \cmark & \cmark & \cmark & \cmark & \cmark & \cmark & \cmark & \cmark & \cmark & \cmark & \cmark & \cmark \\
26474-7 & \cmark & \cmark & \cmark & \cmark & \cmark & \cmark & \cmark & \cmark & \cmark & \cmark & \cmark & \cmark & \cmark & \cmark & \cmark & \cmark & \cmark & \cmark & \cmark & \cmark \\
26478-8 & \cmark & \cmark & \cmark & \cmark & \cmark & \cmark & \cmark & \cmark & \cmark & \cmark & \cmark & \cmark & \cmark & \cmark & \cmark & \cmark & \cmark & \cmark & \cmark & \cmark \\
26484-6 & \cmark & \cmark & \cmark & \cmark & \cmark & \cmark & \cmark & \cmark & \cmark & \cmark & \cmark & \cmark & \cmark & \cmark & \cmark & \cmark & \cmark & \cmark & \cmark & \cmark \\
26485-3 & \cmark & \cmark & \cmark & \cmark & \cmark & \cmark & \cmark & \cmark & \cmark & \cmark & \cmark & \cmark & \cmark & \cmark & \cmark & \cmark & \cmark & \cmark & \cmark & \cmark \\
26499-4 & \cmark & \cmark & \cmark & \cmark & \cmark & \cmark & \cmark & \cmark & \cmark & \cmark & \cmark & \cmark & \cmark & \cmark & \cmark & \cmark & \cmark & \cmark & \cmark & \cmark \\
26505-8 & \cmark & \cmark & \cmark & \cmark & \cmark & \cmark & \cmark & \cmark &  &  &  & \cmark & \cmark & \cmark &  &  &  & \cmark &  & \cmark \\
26507-4 & \cmark & \cmark & \cmark & \cmark & \cmark & \cmark &  & \cmark &  &  &  &  & \cmark &  & \cmark & \cmark &  & \cmark &  &  \\
26508-2 & \cmark &  & \cmark &  &  &  &  &  &  &  &  &  &  &  &  &  &  &  &  &  \\
26511-6 & \cmark & \cmark & \cmark & \cmark & \cmark & \cmark & \cmark & \cmark & \cmark & \cmark & \cmark & \cmark & \cmark & \cmark & \cmark & \cmark & \cmark & \cmark & \cmark & \cmark \\
26515-7 & \cmark & \cmark & \cmark & \cmark & \cmark & \cmark & \cmark & \cmark & \cmark & \cmark & \cmark & \cmark & \cmark & \cmark & \cmark & \cmark & \cmark & \cmark & \cmark & \cmark \\
2823-3 & \cmark & \cmark & \cmark & \cmark & \cmark & \cmark & \cmark & \cmark & \cmark & \cmark & \cmark & \cmark & \cmark & \cmark & \cmark & \cmark & \cmark & \cmark & \cmark & \cmark \\
2857-1 &  &  &  &  &  &  &  &  &  &  &  &  &  &  &  &  &  & \cmark &  & \cmark \\
2862-1 &  &  &  &  &  &  &  &  &  &  & \cmark & \cmark &  & \cmark &  &  &  &  &  &  \\
2885-2 & \cmark & \cmark & \cmark & \cmark & \cmark & \cmark & \cmark & \cmark & \cmark & \cmark & \cmark & \cmark & \cmark & \cmark & \cmark & \cmark & \cmark & \cmark & \cmark & \cmark \\
2888-6 &  &  &  & \cmark &  & \cmark &  &  &  &  &  & \cmark &  & \cmark &  & \cmark &  &  &  & \cmark \\
2889-4 &  &  &  &  &  &  &  &  &  &  &  & \cmark &  &  &  &  &  &  &  &  \\
2947-0 &  & \cmark &  & \cmark &  & \cmark &  & \cmark &  &  & \cmark & \cmark & \cmark & \cmark & \cmark & \cmark &  & \cmark & \cmark & \cmark \\
2951-2 & \cmark & \cmark & \cmark & \cmark & \cmark & \cmark & \cmark & \cmark & \cmark & \cmark & \cmark & \cmark & \cmark & \cmark & \cmark & \cmark & \cmark & \cmark & \cmark & \cmark \\
2986-8 &  &  &  &  &  &  &  &  &  &  &  &  &  &  &  &  &  &  &  & \cmark \\
3016-3 &  & \cmark &  &  &  & \cmark &  & \cmark &  & \cmark & \cmark &  & \cmark & \cmark & \cmark & \cmark & \cmark & \cmark &  &  \\
30180-4 & \cmark & \cmark & \cmark & \cmark & \cmark & \cmark & \cmark & \cmark & \cmark & \cmark & \cmark & \cmark & \cmark & \cmark & \cmark & \cmark & \cmark & \cmark & \cmark & \cmark \\
3024-7 &  &  &  & \cmark &  & \cmark &  &  &  & \cmark & \cmark &  & \cmark & \cmark &  & \cmark &  &  &  & \cmark \\
3026-2 &  &  &  & \cmark &  &  &  &  &  &  & \cmark &  & \cmark &  &  & \cmark & \cmark & \cmark &  &  \\
30376-8 & \cmark &  &  &  &  &  &  &  &  &  &  &  &  &  &  &  &  &  &  &  \\
30394-1 & \cmark & \cmark & \cmark & \cmark & \cmark & \cmark & \cmark & \cmark & \cmark & \cmark & \cmark & \cmark & \cmark & \cmark & \cmark &  &  & \cmark & \cmark & \cmark \\
30395-8 & \cmark & \cmark & \cmark & \cmark & \cmark & \cmark & \cmark & \cmark & \cmark &  & \cmark &  & \cmark & \cmark & \cmark & \cmark & \cmark & \cmark & \cmark & \cmark \\
30451-9 & \cmark &  & \cmark & \cmark &  & \cmark &  & \cmark & \cmark &  & \cmark & \cmark & \cmark & \cmark & \cmark &  &  & \cmark &  &  \\
30522-7 &  & \cmark &  &  &  &  &  &  &  &  &  &  &  &  &  &  &  &  &  & \cmark \\
3084-1 & \cmark &  & \cmark &  & \cmark & \cmark &  & \cmark &  & \cmark & \cmark & \cmark &  & \cmark & \cmark & \cmark &  & \cmark &  & \cmark \\
3094-0 & \cmark & \cmark & \cmark & \cmark & \cmark & \cmark & \cmark & \cmark & \cmark & \cmark & \cmark & \cmark & \cmark & \cmark & \cmark & \cmark & \cmark & \cmark & \cmark & \cmark \\
33358-3 &  &  &  &  &  &  &  &  &  &  &  & \cmark &  &  &  &  &  &  &  &  \\
33558-8 &  & \cmark &  & \cmark &  & \cmark &  &  &  &  &  & \cmark &  &  &  &  &  &  &  &  \\
33944-0 &  &  &  &  &  &  &  &  &  &  &  & \cmark &  &  &  &  &  & \cmark &  &  \\
35591-7 &  & \cmark &  & \cmark &  & \cmark &  &  &  &  &  & \cmark & \cmark & \cmark &  & \cmark &  & \cmark & \cmark & \cmark \\
36916-5 &  &  &  &  &  &  &  &  &  &  &  & \cmark &  &  &  &  &  &  &  &  \\
38176-4 &  &  &  &  &  &  &  &  &  &  &  & \cmark &  &  &  &  &  &  &  &  \\
38483-4 &  &  &  & \cmark &  &  &  &  &  &  & \cmark &  &  & \cmark &  &  &  &  &  & \cmark \\
38875-1 &  & \cmark &  &  &  &  &  &  &  &  &  &  &  & \cmark &  &  &  & \cmark &  &  \\
42483-8 &  &  &  &  &  &  &  &  &  &  &  & \cmark &  &  &  &  &  &  &  &  \\
4537-7 &  & \cmark &  &  &  &  &  &  &  &  &  &  &  &  &  &  &  &  &  &  \\
4544-3 & \cmark & \cmark & \cmark & \cmark & \cmark & \cmark & \cmark & \cmark & \cmark & \cmark & \cmark & \cmark & \cmark & \cmark & \cmark & \cmark & \cmark & \cmark & \cmark & \cmark \\
4548-4 &  & \cmark &  &  &  &  &  &  &  &  &  &  &  & \cmark &  &  &  &  &  &  \\
48642-3 &  & \cmark &  & \cmark & \cmark & \cmark & \cmark & \cmark & \cmark & \cmark & \cmark & \cmark & \cmark & \cmark & \cmark & \cmark & \cmark & \cmark & \cmark & \cmark \\
48643-1 & \cmark & \cmark & \cmark & \cmark & \cmark & \cmark & \cmark & \cmark & \cmark & \cmark & \cmark & \cmark & \cmark & \cmark & \cmark & \cmark & \cmark & \cmark & \cmark & \cmark \\
53962-7 &  &  &  &  &  &  &  &  & \cmark &  &  &  &  &  &  &  &  &  &  &  \\
5902-2 & \cmark & \cmark &  & \cmark & \cmark & \cmark & \cmark & \cmark &  & \cmark & \cmark & \cmark & \cmark & \cmark & \cmark & \cmark & \cmark & \cmark & \cmark & \cmark \\
5905-5 & \cmark & \cmark & \cmark & \cmark & \cmark & \cmark & \cmark & \cmark & \cmark & \cmark & \cmark & \cmark & \cmark & \cmark & \cmark & \cmark & \cmark & \cmark & \cmark & \cmark \\
6298-4 &  & \cmark &  &  &  & \cmark &  &  &  &  &  & \cmark &  & \cmark &  &  &  & \cmark & \cmark & \cmark \\
6690-2 & \cmark & \cmark & \cmark & \cmark & \cmark & \cmark & \cmark & \cmark & \cmark & \cmark & \cmark & \cmark & \cmark & \cmark & \cmark & \cmark & \cmark & \cmark & \cmark & \cmark \\
6768-6 & \cmark & \cmark & \cmark & \cmark & \cmark & \cmark & \cmark & \cmark & \cmark & \cmark & \cmark & \cmark & \cmark & \cmark & \cmark & \cmark & \cmark & \cmark & \cmark & \cmark \\
6875-9 &  & \cmark &  &  &  &  &  &  &  &  &  &  &  &  &  &  &  &  &  &  \\
69405-9 &  & \cmark & \cmark & \cmark &  & \cmark & \cmark & \cmark & \cmark & \cmark & \cmark & \cmark & \cmark & \cmark & \cmark & \cmark & \cmark & \cmark & \cmark & \cmark \\
704-7 & \cmark & \cmark &  & \cmark & \cmark & \cmark & \cmark & \cmark & \cmark & \cmark & \cmark & \cmark & \cmark & \cmark & \cmark & \cmark & \cmark & \cmark & \cmark & \cmark \\
705-4 & \cmark & \cmark &  &  &  &  &  &  &  &  &  &  &  &  &  &  &  &  &  &  \\
706-2 & \cmark & \cmark & \cmark & \cmark & \cmark & \cmark & \cmark & \cmark & \cmark & \cmark & \cmark & \cmark & \cmark & \cmark & \cmark & \cmark & \cmark & \cmark & \cmark & \cmark \\
707-0 & \cmark & \cmark &  &  &  & \cmark &  &  &  &  &  & \cmark & \cmark & \cmark & \cmark &  &  & \cmark & \cmark &  \\
712-0 & \cmark & \cmark &  & \cmark &  &  &  &  &  &  & \cmark & \cmark &  & \cmark & \cmark &  & \cmark &  &  & \cmark \\
714-6 & \cmark & \cmark &  & \cmark & \cmark & \cmark &  & \cmark &  & \cmark &  & \cmark & \cmark & \cmark & \cmark & \cmark & \cmark & \cmark &  & \cmark \\
718-7 & \cmark & \cmark & \cmark & \cmark & \cmark & \cmark & \cmark & \cmark & \cmark & \cmark & \cmark & \cmark & \cmark & \cmark & \cmark & \cmark & \cmark & \cmark & \cmark & \cmark \\
731-0 & \cmark & \cmark & \cmark & \cmark & \cmark & \cmark & \cmark & \cmark & \cmark & \cmark & \cmark & \cmark & \cmark & \cmark & \cmark & \cmark & \cmark & \cmark & \cmark & \cmark \\
732-8 & \cmark & \cmark &  & \cmark &  & \cmark &  & \cmark &  &  & \cmark & \cmark &  &  & \cmark &  & \cmark & \cmark &  & \cmark \\
736-9 & \cmark & \cmark & \cmark & \cmark & \cmark & \cmark & \cmark & \cmark & \cmark & \cmark & \cmark & \cmark & \cmark & \cmark & \cmark & \cmark & \cmark & \cmark & \cmark & \cmark \\
737-7 & \cmark & \cmark &  & \cmark &  & \cmark &  & \cmark &  &  & \cmark & \cmark & \cmark & \cmark &  &  & \cmark & \cmark &  & \cmark \\
742-7 & \cmark & \cmark &  & \cmark & \cmark & \cmark & \cmark & \cmark & \cmark & \cmark & \cmark & \cmark & \cmark & \cmark & \cmark & \cmark & \cmark & \cmark & \cmark & \cmark \\
743-5 & \cmark & \cmark & \cmark & \cmark &  & \cmark &  & \cmark &  &  &  &  &  &  &  &  &  & \cmark &  & \cmark \\
744-3 & \cmark & \cmark & \cmark & \cmark &  & \cmark &  &  & \cmark &  & \cmark & \cmark & \cmark & \cmark &  & \cmark &  & \cmark & \cmark & \cmark \\
751-8 & \cmark & \cmark &  & \cmark & \cmark & \cmark & \cmark & \cmark & \cmark & \cmark & \cmark & \cmark & \cmark & \cmark & \cmark & \cmark & \cmark & \cmark & \cmark & \cmark \\
753-4 & \cmark &  &  &  &  &  &  &  &  &  &  &  &  &  &  &  &  &  &  &  \\
764-1 &  & \cmark &  &  &  & \cmark &  &  &  &  &  &  & \cmark & \cmark &  &  &  & \cmark &  &  \\
769-0 & \cmark & \cmark &  & \cmark &  &  &  &  &  &  &  &  &  &  &  &  &  & \cmark &  &  \\
770-8 &  & \cmark &  & \cmark &  & \cmark &  &  &  &  & \cmark & \cmark & \cmark & \cmark & \cmark &  &  & \cmark & \cmark & \cmark \\
77146-9 &  &  &  &  &  &  &  &  &  &  &  &  &  &  &  &  &  & \cmark &  &  \\
777-3 & \cmark & \cmark & \cmark & \cmark & \cmark & \cmark &  & \cmark & \cmark & \cmark & \cmark & \cmark & \cmark & \cmark & \cmark & \cmark & \cmark & \cmark & \cmark & \cmark \\
789-8 & \cmark & \cmark & \cmark & \cmark & \cmark & \cmark &  & \cmark &  & \cmark & \cmark & \cmark & \cmark & \cmark & \cmark & \cmark &  & \cmark & \cmark & \cmark \\
98979-8 &  & \cmark &  & \cmark &  &  &  &  &  &  & \cmark &  &  & \cmark & \cmark &  &  &  &  & \cmark \\
\end{longtable}
\normalsize

\newpage
\section{Extended Results}
\label{app:results}

\subsection{Sample Size Ablation (GDT vs Baselines on 2k Patients)}
\label{app:forecast_2k_section}

\begin{figure}[h!]
    \centering
    \includegraphics[width=\linewidth]{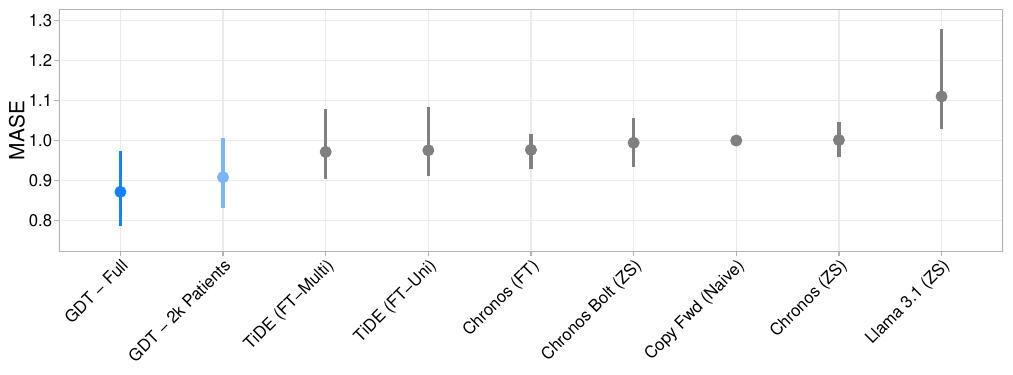}
    \caption{\textbf{GDT trained on the same 2,000-patient subset as the baselines still outperforms them, though it underperforms the full model.} Comparison of MASE for forecasting blood variables. The reduced-data GDT (Light Blue) surpasses TiDE and Chronos but does not match the performance of the GDT model trained on the full pan-cancer dataset (Teal), highlighting the benefits of scale. ZS denotes zero-shot, FT is fine tuned, Uni is univariate input, Multi is multivariate input.}
    \label{fig:app:forecast_2k}
\end{figure}

To ensure a fair comparison, we run GDT also trained on the identical samples as the baselines, seeing that it outperforms the models, yet is worse than the model trained on the full dataset, seen in Figure \ref{fig:app:forecast_2k}.

\newpage
\subsection{Forecasting Full Results (RWD)}
\label{app:forecasting_full}

\begin{figure}[h!]
    \centering
    \includegraphics[width=\linewidth]{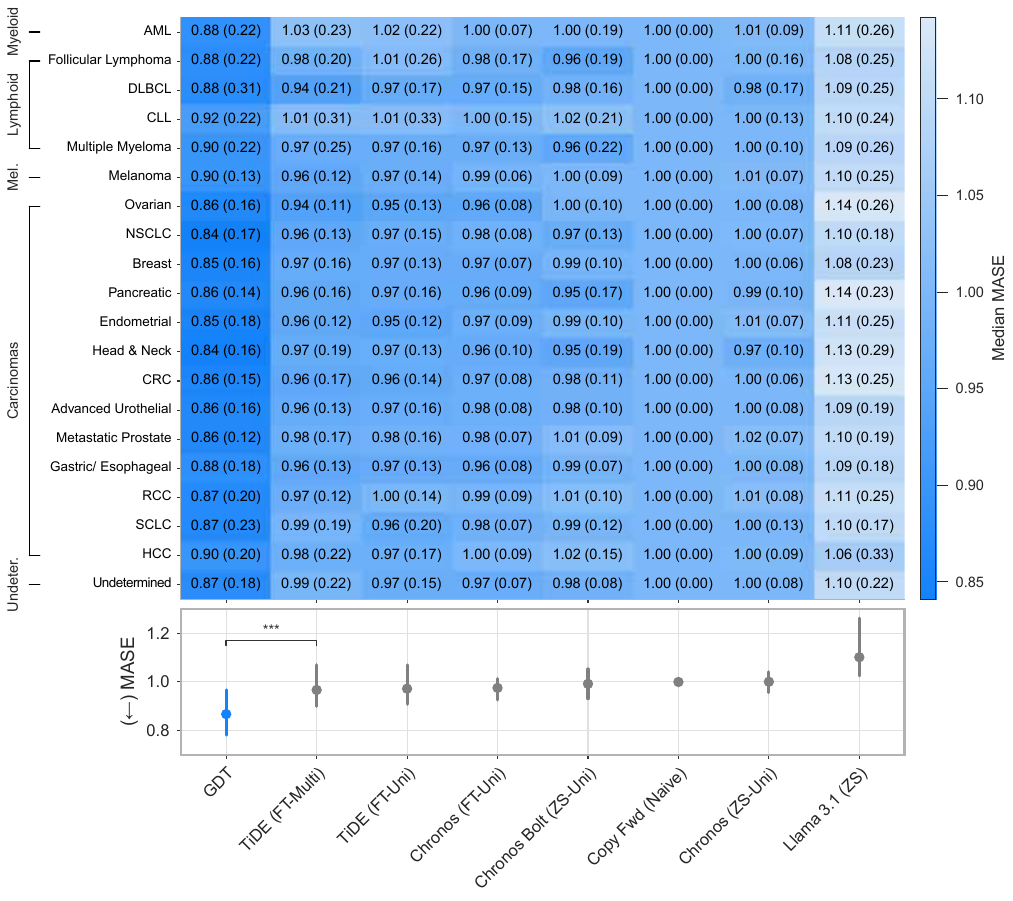}
    \caption{\textbf{Comprehensive blood biomarker forecasting results across all variables and indications.} The heatmap displays the Median MASE for GDT versus baselines across all 20 cancer types. GDT consistently achieves lower error rates across both myeloid/lymphoid and solid tumor indications compared to standard time-series models and zero-shot LLMs. ZS denotes zero-shot, FT is fine tuned, Uni is univariate input, Multi is multivariate input.}
    \label{fig:app:forecasting_full}
\end{figure}

We present the forecasting results across all variables in Figure \ref{fig:app:forecasting_full}.

\subsection{Clinical Trial Task Setup}
\label{app:setup:clinical_trials}

To construct the dataset, we prioritized the largest combination of clinical trials featuring overlapping treatment regimens. The two largest trials were designated as the training set for the experimental settings with fine-tuning: OAK ($n = 1126$; comparing atezolizumab vs. docetaxel) and IMpower131 ($n = 949$; comparing atezolizumab + carboplatin + paclitaxel vs. atezolizumab + carboplatin + nab-paclitaxel vs. carboplatin + nab-paclitaxel). For internal validation and hyperparameter tuning, we employed a random 20\% holdout subset of these training trials. To rigorously evaluate the model's ability to generalize across different patient populations and protocols, we adopted a leave-trial-out testing strategy rather than a random pooled split. We utilized the two largest trials (OAK, IMpower131) for training to maximize data diversity during learning, while reserving two distinct, independent trials (POPLAR, IMpower130) strictly for external testing: POPLAR ($n = 263$; atezolizumab vs. docetaxel) and IMpower130 ($n = 680$; atezolizumab + carboplatin + nab-paclitaxel vs. carboplatin + nab-paclitaxel). We used 17 blood biomarkers for forecasting, as well as survival and progression for event evaluation.

\subsection{Further Reasoning Results}
\label{app:reasoning:results}

\subsubsection{Reasoning Details}

To validate the clinical semantic alignment of the generated rationales, we employed a keyword-based concept analysis focusing on four distinct therapy regimens found in the NSCLC cohort: Chemotherapy, Immunotherapy (IO), Targeted Therapy, and Chemo-IO combinations. We manually curated a dictionary of inclusion and exclusion terms for specific clinical concepts (e.g., `Marrow Suppression', `Immune Activation', `EGFR mutation'). Inclusion terms captured synonymous phrases and specific biological mechanisms, while exclusion terms were strictly applied to filter out negated contexts or unrelated mentions to ensure false positives were minimized.

For each concept within a therapy group, we computed two metrics:
\begin{itemize}
    \item \textbf{Frequency:} The percentage of generated reasoning chains within a specific therapy group that contained at least one verified instance of the concept.
    \item \textbf{Relative Importance:} A normalized measure of concept specificity calculated row-wise. For a given concept, this is defined as its frequency in the specific therapy group divided by the maximum frequency of that concept observed across all four therapy groups. This scaling ensures that the therapy group with the strongest association for a given concept is assigned a value of 1.0, highlighting relative specificity regardless of absolute frequency.
\end{itemize}

\subsubsection{Reasoning Embeddings Visualization}
To assess the reasoning chains, we extract them from the test set, and then embed them using Qwen3-8B. Finally, we perform UMAP and two-dimensional display of the embedded data.

\begin{figure*}[h!]
    \centering
    \includegraphics[width=\linewidth]{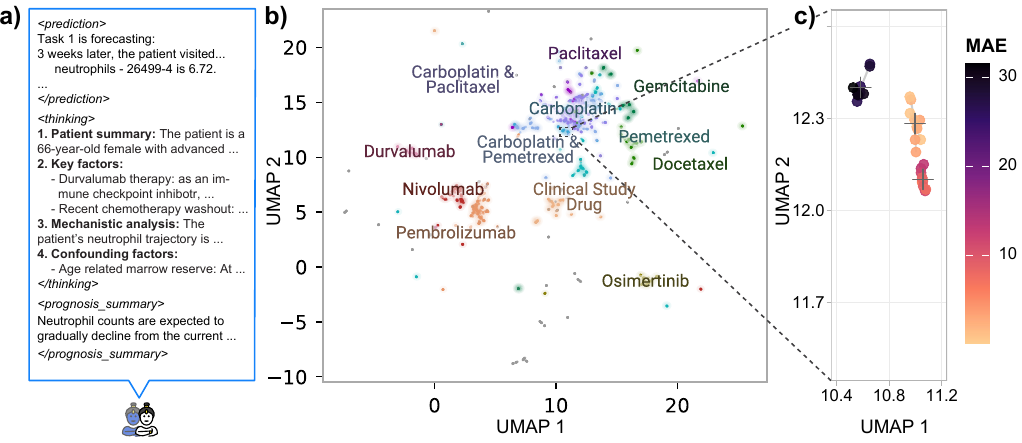}
    \caption{\textbf{GDT can be extended to provide reasoning, exemplified here in NSCLC neutrophil forecasting.} \textbf{a)} Example of a generated prediction and reasoning chain. \textbf{b)} UMAP visualization of reasoning chain embeddings, colored by the top 12 most common categories, shows structural alignment with clinical factors such as therapy types. \textbf{c)} Zoomed view of the patient with the highest MAE on the test set; the centroids (+) and clustering indicate that generated reasoning chains are consistent across multiple runs.}
\end{figure*}

\subsubsection{Example Reasoning Chain}

\begin{modelbox}
<prediction>
Task 1 is forecasting:
3 weeks later, the patient visited and experienced the following: 
	neutrophils - 26499-4 is 3.4.

2 weeks later, the patient visited and experienced the following: 
	neutrophils - 26499-4 is 4.6.

2 weeks later, the patient visited and experienced the following: 
	neutrophils - 26499-4 is 4.8.

2 weeks later, the patient visited and experienced the following: 
	neutrophils - 26499-4 is 4.6.

2 weeks later, the patient visited and experienced the following: 
	neutrophils - 26499-4 is 3.9.

</prediction>
<thinking>
1. **Patient Summary:** The patient is a 52-year-old male with metastatic NSCLC, previously diagnosed with prostate cancer, currently on nivolumab (first-line immune checkpoint inhibitor) as his most recent therapy. He has progressive disease with bone and liver metastases, confirmed PD-L1 negativity on both tumor and immune cells, and a complex genomic profile including TP53, STK11, SMARCA4, and CDKN2A mutations, with MSI-stable status and no actionable EGFR or ALK alterations. Recent labs show stable renal function, mild anemia, and fluctuating but generally preserved neutrophil counts (last value: 4.4 x 10^9/L). ECOG has improved from 2 to 0, indicating clinical stabilization. No active infection or acute inflammatory events are documented. The patient is not receiving G-CSF or other myeloid growth factors.

2. **Key Predictive Factors:**
   - **Nivolumab monotherapy:** Immune checkpoint inhibitors like nivolumab rarely cause severe neutropenia but can induce immune-mediated bone marrow suppression or transient cytopenias via T-cell activation against hematopoietic precursors. Neutrophil counts typically remain stable or show mild fluctuations.
   - **Recent partial response (PR) to therapy:** The documented PR suggests effective tumor control, which reduces tumor-derived cytokines (e.g., IL-6, G-CSF) that can stimulate neutrophil production. This may lead to a gradual normalization of neutrophil counts toward baseline.
   - **Persistent bone metastases:** Bone marrow infiltration by metastatic NSCLC may impair hematopoietic reserve, particularly affecting granulopoiesis. However, the absence of severe cytopenias (e.g., hemoglobin >13, platelets >300) suggests compensatory hematopoiesis is still functional.
   - **Genomic profile (TP53, STK11, SMARCA4 mutations):** These mutations are associated with genomic instability and may impair DNA repair in hematopoietic stem cells, potentially increasing susceptibility to subclinical marrow stress or delayed recovery after minor insults.
   - **Trend of neutrophil stability over prior visits:** Neutrophils have fluctuated between 3.5-5.2 over the last 5 visits (spanning 10 weeks), with no consistent downward trend. This suggests a homeostatic equilibrium under current conditions.

3. **Mechanistic Analysis:** The neutrophil trajectory is governed by a balance between immune-mediated stress and compensatory hematopoiesis. Nivolumab does not directly suppress neutrophil production but may induce low-grade immune activation that occasionally triggers transient neutrophil consumption or redistribution. The partial response reduces tumor-derived inflammatory signals that previously stimulated neutrophil release, leading to a gradual normalization of baseline neutrophil production. However, the presence of bone metastases limits the marrow's capacity for robust rebound, preventing overshoot. The TP53/STK11/SMARCA4 mutations may slightly delay recovery from minor stressors (e.g., subclinical inflammation or transient immune activation), explaining why neutrophil counts do not fully return to pre-treatment levels. The observed fluctuations (e.g., 4.4 -> 3.4 -> 4.6 -> 4.8 -> 4.6 -> 3.9) reflect this dynamic equilibrium: transient immune activation (possibly from tumor cell death or inflammatory signaling from the tumor microenvironment) causes a slight dip, followed by compensatory granulopoiesis. The 2-week intervals between visits allow for partial recovery before the next immune-mediated perturbation, resulting in a cyclical pattern without sustained decline. The absence of G-CSF or infection means no external stimulus for rapid rise or crash.

4. **Confounding Factors:** Subclinical infection (e.g., from oral soft tissue disease or lymph node enlargement) could trigger neutrophilia, but no fever or elevated CRP/ESR is documented. Advanced age (52) is not a major risk factor for neutropenia in this context, but reduced marrow reserve due to prior cancer and metastases may blunt recovery if stressors accumulate.
</thinking>

<prognosis_summary>
Neutrophil counts are expected to remain within the low-normal to normal range with minor fluctuations, reflecting a dynamic equilibrium between immune-mediated stress and compensatory hematopoiesis, with no evidence of progressive myelosuppression. The primary clinical risk is transient neutropenia without febrile episodes, unlikely to require treatment interruption.
</prognosis_summary>
\end{modelbox}

\subsection{Single Indication Training vs Pan-Cancer}
\label{app:forecast_single_indication_overview}

To understand the contribution of the pan-cancer training, we performed an ablation study where we trained GDT on the three cancer types with the fewest patients using only cancer indication data. We focus on the fewest patient counts, as they are most likely to benefit from the pan-cancer capabilities and transfer learning.

\begin{figure}[h!]
    \centering
    \includegraphics[width=0.8\linewidth]{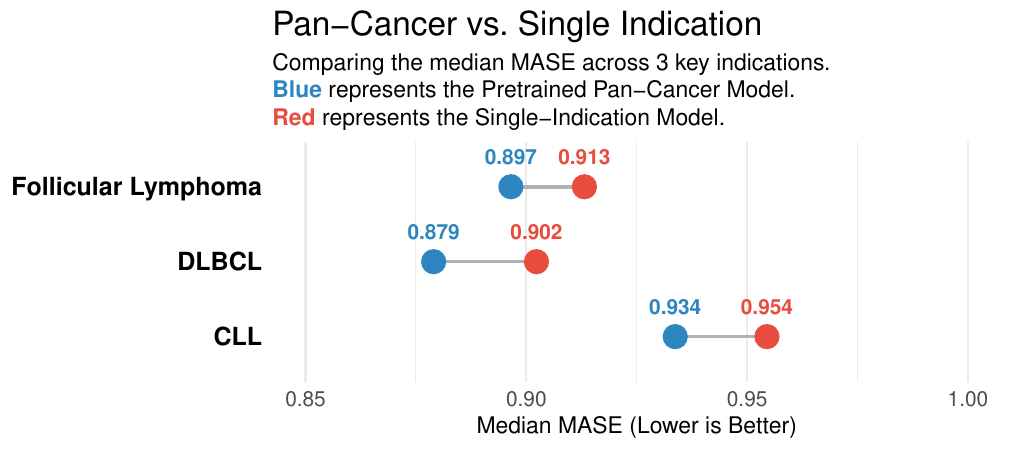}
    \caption{\textbf{Pan-cancer training consistently improves performance over single-indication models.} Comparison of Median MASE for the three cancer indications with the fewest patients ($<500$). The Pan-cancer GDT model (Blue) achieves lower error rates than models trained solely on the specific indication data (Red), demonstrating the benefits of transfer learning in low-data regimes.}
    \label{fig:app:forecast_single_indication}
\end{figure}

In Figure \ref{fig:app:forecast_single_indication}, we see that the pan-cancer model consistently improves performance over the single indication models, albeit at a small level. 

\newpage
\subsection{Assessing Longitudinal Input Length on Performance}
\label{app:ablation_input}

\begin{figure}[h!]
    \centering
    \includegraphics[width=0.4\linewidth]{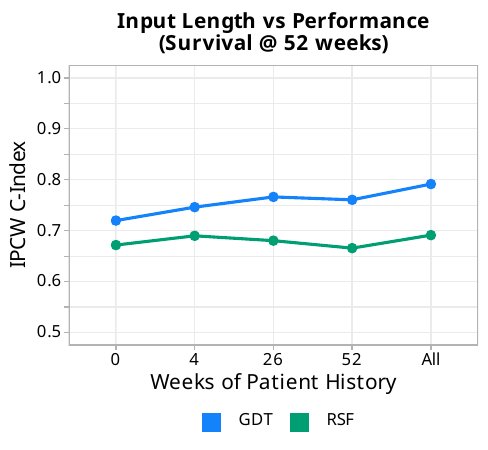}
    \caption{\textbf{Longer historic patient trajectories improve GDT, implying that clinical trials could benefit from additional information on patient history.} Here, we compare Random Survival Forest (RSF) with GDT on NSCLC for 52-week survival prediction on IPCW C-index. We select patients from the test set, patients with at least 52 weeks of input history, then split only at the start of therapy, and systematically cut the input history to the specific input length. 0-4 weeks can be compared to the data currently commonly gathered in clinical trials. We see that more input history improves model performance, implying that clinical trials could benefit from longer patient timelines. We additionally see that GDT outperforms RSF consistently, which focuses on last observed values only.}
    \label{fig:app:ablation_input}
\end{figure}

\subsection{Detailed Clinical Trial Forecasting Results}
\label{app:results:clinical_trial_forecasting}

We provide the comprehensive breakdown of forecasting performance across all models for the clinical trial validation. On the trial POPLAR, the performance ranking by median MASE is as follows: GDT SFT (0.883, IQR 0.09), TiDE fine-tuned univariate (0.936, IQR 0.07), GDT zero-shot (0.947, IQR 0.121), Chronos fine-tuned (0.956, IQR 0.054), TiDE fine-tuned multivariate (0.993, IQR 0.131), Chronos zero-shot (0.998, IQR 0.001), Copy Forward (1.0), Chronos Bolt zero-shot (1.0), and Llama 3.1 (1.164, IQR 0.175).

For trial IMpower130, the ranking follows a similar pattern: GDT SFT (0.754, IQR 0.385), TiDE fine-tuned univariate (0.840, IQR 0.119), GDT zero-shot (0.874, IQR 0.163), Chronos fine-tuned (0.885, IQR 0.075), TiDE fine-tuned multivariate (0.853, IQR 0.208), Chronos zero-shot (0.998, IQR 0.001), Copy Forward (1.0), Chronos Bolt zero-shot (1.0), and Llama 3.1 (1.126, IQR 0.156).

We observe that the TiDE univariate model generally outperforms the multivariate equivalent in this setting. We hypothesize that the reason is the focus on baseline information, limiting the correlations the multivariate model can capture, thereby allowing the univariate model to converge more easily.

\newpage
\subsection{Isotonic Regression Results}
\label{app:results:isotonic}

\subsubsection{Pan-Cancer Real-World Data}

\begin{figure}[h!]
    \centering
    \includegraphics[width=\linewidth]{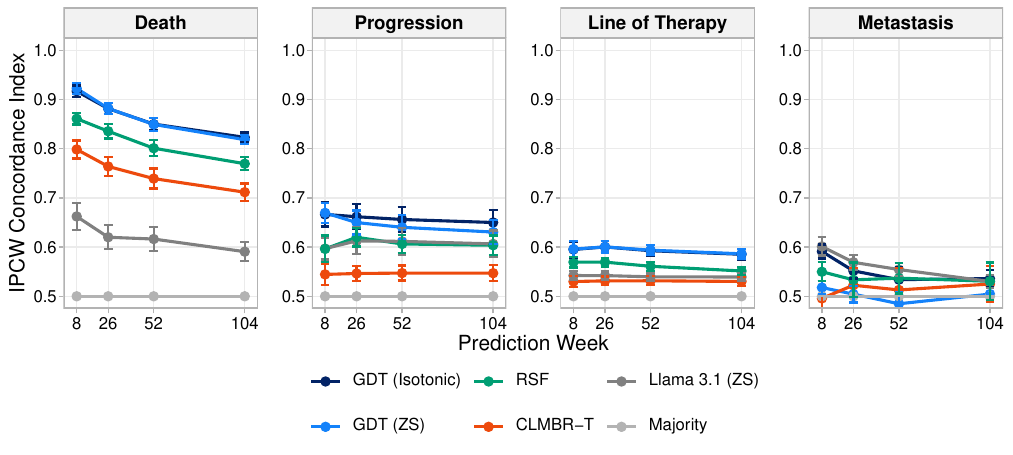}
    \caption{\textbf{Isotonic regression post-processing maintains robust ranking performance on pan-cancer real-world data.} The plots show the IPCW C-Index for Survival and Disease Progression. The performance of the isotonic version (Teal) is highly similar to the original GDT outputs (Light Blue), indicating that enforcing monotonicity does not degrade the model's discriminative ability.}
    \label{fig:app:isotnic:cgdb:c_index}
\end{figure}

\begin{figure}[h!]
    \centering
    \includegraphics[width=\linewidth]{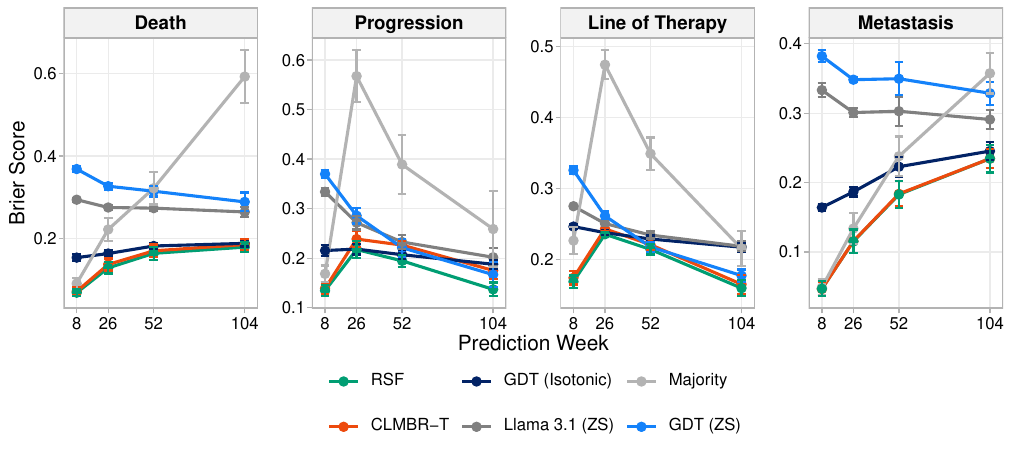}
   \caption{\textbf{Brier score analysis shows improvement over the base model with isotonic regression but indicates a need for further calibration.} Lower Brier scores indicate better calibration. While isotonic regression improves the score compared to the raw GDT output, particularly for Line of Therapy and Death, the results suggest that further work is needed to generate well-calibrated probabilities.}
    \label{fig:app:isotnic:cgdb:brier}
\end{figure}

In the pan-cancer real-world data results, we see an overall slight improvement when including isotonic regression (Figure \ref{fig:app:isotnic:cgdb:c_index}). The Brier score shows an increased performance over the base model. However, further work is required for well calibrated probabilities, as seen in Figure \ref{fig:app:isotnic:cgdb:brier}.

\newpage
\subsubsection{Clinical Trial Results}

\begin{figure}[h!]
    \centering
    \includegraphics[width=\linewidth]{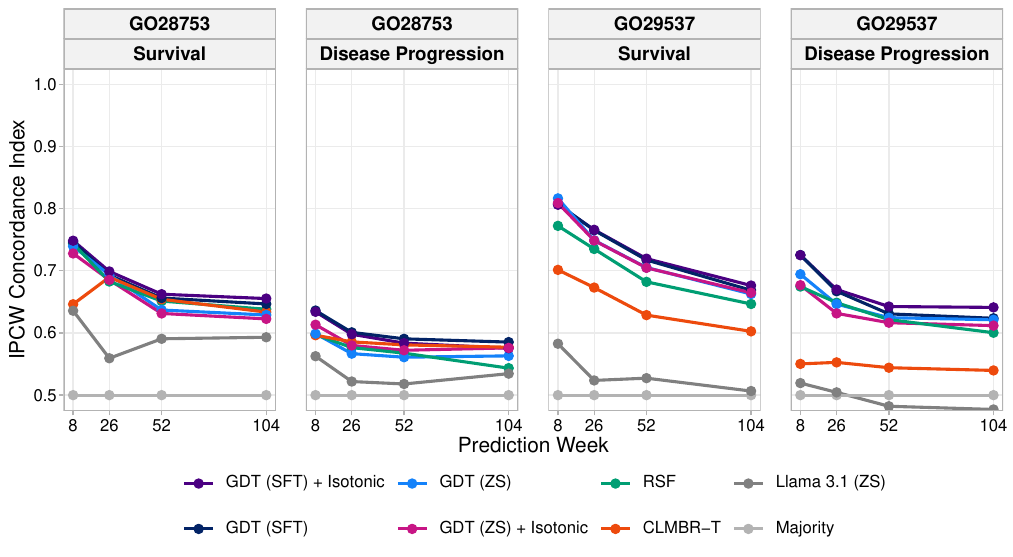}
    \caption{\textbf{Isotonic regression has minimal impact on clinical trial C-Index as GDT predictions are largely naturally monotonic.} Results for trials POPLAR (GO28753) and IMpower130 (GO29537) show that the isotonic regression curves overlap significantly with the standard GDT curves, as only a small fraction (0.37\%) of the zero-shot samples violated monotonicity.}
    \label{fig:app:isotnic:cit:c_index}
\end{figure}

\begin{figure}[h!tb]
    \centering
    \includegraphics[width=\linewidth]{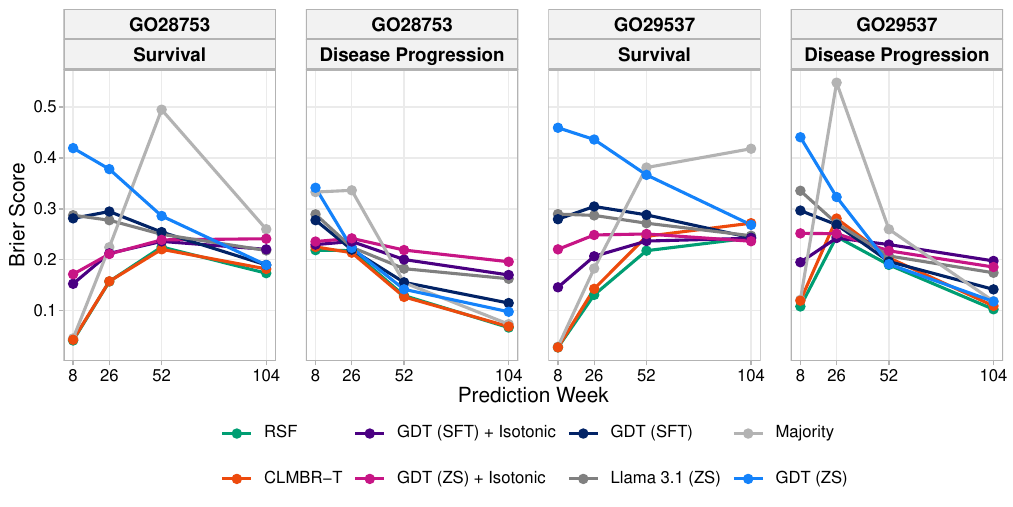}
    \caption{\textbf{Brier scores for clinical trials highlight the need for improved probability calibration despite strong ranking performance.} Comparing the Brier scores across models on trials POPLAR (GO28753) and IMpower130 (GO29537) reveals that while GDT ranks patients effectively (high C-Index), the absolute probability estimates require refinement to minimize the Brier score.}
    \label{fig:app:isotnic:cit:brier}
\end{figure}


\newpage
\subsection{Example Forecast Predictions}

\begin{figure}[h!]
    \centering
    \includegraphics[width=\linewidth]{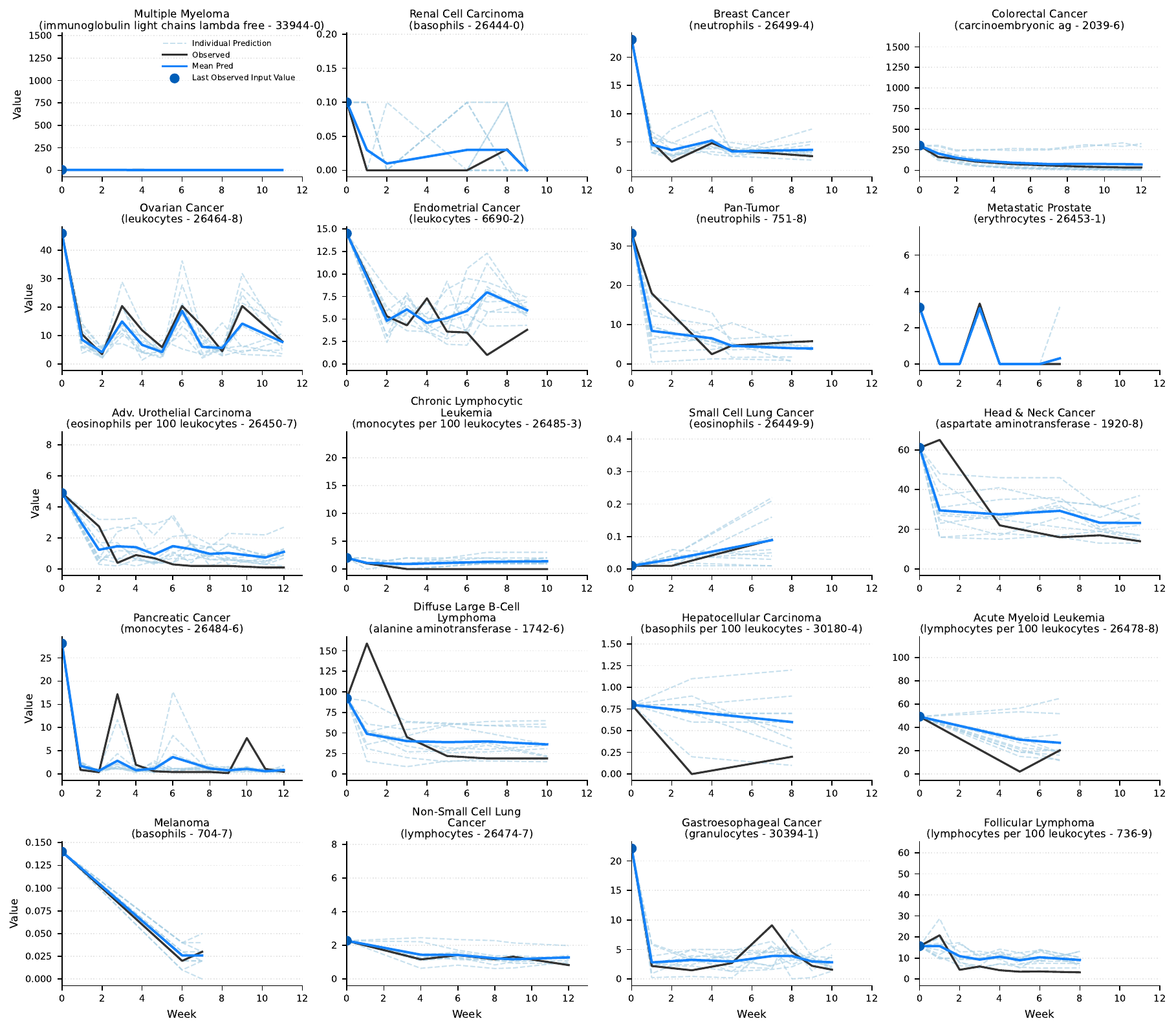}
    \caption{\textbf{Example blood biomarker forecasts by GDT.} The longitudinal trajectory of different blood-based markers are shown for a variety of tumor indications. Measured values are displayed in grey and predicted values in blue. Note that in some cases where the average prediction is incorrect, individual predictions sometimes are still able to capture the shape or values, such as the first spike in pancreatic cancer, or the dip in hepatocellular carcinoma. However, in other cases GDT is not able to capture large variability in the results, such as an initial strong increase in the alanine aminotransferase values in DLBCL.}
    \label{fig:app:forecast_examples}
\end{figure}

\newpage
\subsection{Synthetic Patient Example}

Using a synthetic patient to ensure data privacy compliance.

Input:

\begin{modelbox}

The following is a patient, starting with the demographic data, following visit by visit everything that the patient experienced. All lab codes refer to LOINC codes.

Starting with demographic data:
	Patient gender is male,
	age of patient is 77 years,
    ...

On the first visit, the patient experienced the following: 
	advanced cancer diagnosis is non small cell lung cancer,
	initial cancer diagnosis is non small cell lung cancer.
	metastasis Pleura is diagnosed.

2 weeks later, the patient visited and experienced the following: 
	<genetic>
	...
	Tissue:  pleura Gene: CHEK2 short variant: frameshift parameters:  likely truncation,
	...
	</genetic>.


1 weeks later, the patient visited and experienced the following: 
	ECOG is 2,
	alanine aminotransferase - 1742-6 is 9,
	albumin - 1751-7 is 39,
	calcium - 17861-6 is 8.9,
	aspartate aminotransferase - 1920-8 is 20,
	bilirubin - 1975-2 is 0.3,
	carbon dioxide - 2028-9 is 25,
	neutrophils per 100 leukocytes - 26511-6 is 72,
	platelets - 26515-7 is 419,
	...
	potassium - 2823-3 is 4.7,
	protein - 2885-2 is 61,
	sodium - 2951-2 is 132,
	basophils per 100 leukocytes - 30180-4 is 0.3,
	...


1 weeks later, the patient visited and experienced the following: 
	Other fatigue is diagnosed,
	Encounter for antineoplastic immunotherapy - Z51.12 is diagnosed,
	drug pembrolizumab is 200,
	ECOG is 1,
	alanine aminotransferase - 1742-6 is 9,
	albumin - 1751-7 is 42,
	calcium - 17861-6 is 9.4,
	aspartate aminotransferase - 1920-8 is 21,
	neutrophils - 26499-4 is 5.9,
	neutrophils per 100 leukocytes - 26511-6 is 73.2,
	platelets - 26515-7 is 243,
	...
	thyroxine - 3026-2 is 6.7,
	urea nitrogen - 3094-0 is 10,
	thyroxine free index - 32215-6 is 2,
	line number is 1,
	body height is 180.34,
	body weight is 63.45752,
	oxygen saturation in arterial blood by pulse oximetry is 98.

...



Here we repeat the last observed values of each genetic event in the input data:

0 weeks later, the patient visited and experienced the following: 
	...
	Tissue:  pleura Gene: CHEK2 short variant: frameshift parameters:  likely truncation,
	...
	</genetic>.
	... 

The most recent line of therapy:
	Pembrolizumab

The last values of the variables in the input data are:
	hematocrit - 20570-8 was 33.6
	creatinine - 2160-0 was 0.6
	platelets - 26515-7 was 257

You will now have multiple tasks to complete. Please answer for each task in the same order as they are presented. Before every response state the task nr, e.g. 'Task 2:'.

Task 1 is forecasting:
Your task is to predict the future values of the following variables for each cumulative week starting from the last visit:

	hematocrit - 20570-8 the future weeks 1, 4, 7, 10
	creatinine - 2160-0 the future weeks 1, 4, 7, 10
	platelets - 26515-7 the future weeks 1, 4, 7, 10


Task 2 is time to event prediction:
Your task is to predict whether the following event was censored 1 weeks from the last clinical visit and whether the event occurred or not: death.
Please provide your prediction in the following format: 'Here is the prediction: the event (<name of event>) was [not] censored and [did not occur]/[occurred].'
\end{modelbox}

Output:
\begin{modelbox}
Task 1 is forecasting:
1 weeks later, the patient visited and experienced the following: 
	hematocrit - 20570-8 is 36.1,
	creatinine - 2160-0 is 0.65,
	platelets - 26515-7 is 260.

3 weeks later, the patient visited and experienced the following: 
	hematocrit - 20570-8 is 39,
	creatinine - 2160-0 is 0.6,
	platelets - 26515-7 is 257.

3 weeks later, the patient visited and experienced the following: 
	hematocrit - 20570-8 is 36.6,
	creatinine - 2160-0 is 0.63,
	platelets - 26515-7 is 271.

3 weeks later, the patient visited and experienced the following: 
	hematocrit - 20570-8 is 35.7,
	creatinine - 2160-0 is 0.61,
	platelets - 26515-7 is 257.



Task 2 is time to event prediction:
Here is the prediction: the event (death) was not censored and did not occur.
\end{modelbox}

\subsection{System Prompt}

\begin{modelbox}
As a specialist predictive model in personalized medicine, your task is to forecast the health trajectory of cancer patients by integrating genomic data, lifestyle factors, treatment history and anything else provided about the patient. Use the provided patient data, including genetic mutations, biomarker levels, and previous treatment responses, to predict all requested tasks. Deliver precise and clinically relevant predictions to enhance patient care and treatment planning.
\end{modelbox}

\subsection{Qwen3 Next 80B-A3B Prompt}

The following prompt was used to generate the initial reasoning chains using the Qwen3 Next 80B-A3B model. Note, that we then replace the prediction with the true prediction, and also shift it to the beginning, since this showed to provide more stable results in early experiments.

\begin{modelbox}
You are an expert hematologist-oncologist. You will receive a complete 
patient history and a specific task to predict the patient's neutrophil 
trajectory.

Your primary goal is to generate a step-by-step reasoning chain that 
leads to your prediction. This rationale is more important than the 
prediction itself.

Structure your entire response using the following tags. Do not include 
any text outside of these tags.

<thinking>
Inside this tag, you must follow this four-step reasoning process:

1.  **Patient Summary:** Briefly summarize the patient's current oncological 
    and hematological status. Focus on the diagnosis, active treatments, 
    and the most recent relevant lab values.
2.  **Key Predictive Factors:** Identify the **top 5 most influential factors** from the patient's record that will drive the neutrophil trajectory. 
    List each factor (e.g., specific drug, time since last treatment, 
    comorbidity, recent lab trend) and provide a concise justification 
    for its high importance.
3.  **Mechanistic Analysis:** This is the most critical step. Synthesize 
    the 5 factors you identified. Provide a detailed, step-by-step 
    biological explanation of how these factors will interact to 
    influence the neutrophil count *over time*.
    * Describe the specific biological pathways involved (e.g., 
        myelosuppression from a specific drug class, hematopoietic 
        recovery kinetics, effects of G-CSF on bone marrow 
        precursors, inflammatory cytokine release).
    * Explain the expected *timing* of these effects (e.g., "The 
        patient is X days post-[Chemo], so we expect the nadir 
        around day Y," or "The recent G-CSF administration will 
        likely cause a transient leukocytosis followed by...").
4.  **Confounding Factors:** Briefly mention 1-2 other factors (e.g., 
    potential infection, patient age, nutritional status) that could 
    complicate or alter your primary predicted trajectory.

</thinking>

<prognosis_summary>
Based on your thinking and rationale, provide a 1-2 sentence summary of the expected 
neutrophil trend (e.g., "Expect sharp decline into severe neutropenia," 
or "Anticipate slow but steady recovery") and the primary clinical risk 
(e.g., "High risk of febrile neutropenia," or "Risk of treatment delay").
</prognosis_summary>

<prediction>
[Place the final, formatted prediction here as specified in the task 
description.]
</prediction>
\end{modelbox}


\end{document}